\documentclass[twoside]{article}

\usepackage[accepted]{aistats2025}

\usepackage{amsmath, amssymb, amsthm}
\usepackage{graphicx}
\usepackage{xcolor}  
\usepackage{colortbl}
\usepackage[round]{natbib}

\usepackage[breaklinks=true, colorlinks=true, linkcolor=black, citecolor=black, urlcolor=blue]{hyperref}
\usepackage{xurl}

\usepackage{cleveref}
\usepackage{algpseudocode}
\usepackage[noline,ruled,noend]{algorithm2e}

\newcommand{\marginalplain}{\pi}
\newcommand{\measurefamily}{\mathcal{D}}
\newcommand{\marginal}[1]{\marginalplain_{#1}}

\newcommand{\timet}{t}
\newcommand{\timeidx}{i}
\newcommand{\sampleidx}{n}

\newcommand{\totalnumparticles}{N}
\newcommand{\totsteps}{I}
\newcommand{\timestep}[1]{\timet_{#1}}

\newcommand{\discstep}{\ell}
\newcommand{\totdiscstep}{L}

\newcommand{\bridgeplain}{q}
\newcommand{\bridge}{\bridgeplain}
\newcommand{\popmarginalat}[1]{\tilde{\bridgeplain}_{#1}}

\newcommand{\drift}{{\bm{b}}}
\newcommand{\modeldriftbackward}{\drift_{\text{back}}}

\newcommand{\refdrift}{\drift_{\text{ref}}}

\newcommand{\estimatedriftstep}[1]{\hat{\drift}^{#1}}
\newcommand{\estimaterefdriftstep}[1]{\hat{\drift}_{\text{ref}}^{#1}}
\newcommand{\estimatebackwarddriftstep}[1]{\hat{\drift}_{\text{back}}^{#1}}

\newcommand{\estimatedrift}{\hat{\drift}}
\newcommand{\estimatebkwdrift}{\hat{\drift}_{\text{back}}}

\newcommand{\modelfamily}{\mathcal{F}}

\newcommand{\de}{\text{d}}
\newcommand{\responseplain}{X}
\newcommand{\obsplain}{Y}

\newcommand{\responseat}[1]{\bm{\responseplain}_{#1}}
\newcommand{\simulatedresponseat}[1]{\bm{\hat{\responseplain}}_{#1}}

\newcommand{\obsat}[1]{\bm{\obsplain}_{#1}}
\newcommand{\obsatall}[1]{\bm{\obsplain}_{#1}^{\text{all}}}

\newcommand{\tottrajec}{N}

\newcommand{\brownianm}{\bm{W}_{\timet}}

\newcommand{\volatility}{\gamma}

\newcommand{\modeldrift}{\drift}

\newcommand{\altmeasurefamily}{\mathcal{G}}
\newcommand{\altmodelfamily}{\mathcal{H}}

\newcommand{\refden}{p}

\newcommand{\basemeasure}{\mu}
\newcommand{\basediff}{\de\mu}
\crefname{assumption}{assumption}{assumptions}

\newcommand{\vanilla}{Vanilla-SB}
\newcommand{\dmsbname}{DM-SB}

\newcommand{\vanillaonetime}{Vanilla-SB: one time}
\newcommand{\oursonetime}{Ours: one time}
\newcommand{\oursalltime}{Ours: all times}
\newcommand{\vanillaalltime}{Vanilla-SB: all times}

\newcommand{\dmsb}{DM-SB: one time}
\newcommand{\trajnet}{TrajectoryNet: one time}

\newcommand{\fpm}{$\pm$}

\usepackage{amsmath,amsfonts,bm}
\usepackage{tabularx}

\def\eqref#1{equation~\ref{#1}}

\def\1{\bm{1}}

\DeclareMathAlphabet{\mathsfit}{\encodingdefault}{\sfdefault}{m}{sl}
\SetMathAlphabet{\mathsfit}{bold}{\encodingdefault}{\sfdefault}{bx}{n}

\def\gN{{\mathcal{N}}}

\newcommand{\E}{\mathbb{E}}

\newcommand{\R}{\mathbb{R}}

\newcommand{\KL}{D_{\mathrm{KL}}}

\newcommand{\iteralgo}{k}

\DeclareMathOperator*{\argmax}{arg\,max}
\DeclareMathOperator*{\argmin}{arg\,min}

\usepackage{thmtools}
\usepackage{thm-restate}

\newtheorem{myTheorem}{Theorem}[section]
\newtheorem{myLemma}{Lemma}[section]

\newtheorem{assumption}{Assumption}

\usepackage{booktabs}
\usepackage{multirow}
\begin{document}

%

%

\twocolumn[

\aistatstitle{Multi-marginal Schrödinger Bridges with Iterative Reference Refinement}

\aistatsauthor{ Yunyi Shen$^*$ \And Renato Berlinghieri$^*$ \And  Tamara Broderick }

\aistatsaddress{ MIT \And  MIT \And MIT } ]

\begin{abstract}
  Practitioners often aim to infer an unobserved population trajectory using sample snapshots at multiple time points. E.g. given single-cell sequencing data, scientists would like to learn how gene expression changes over a cell’s life cycle. But sequencing any cell destroys that cell. 
So we can access data for any particular cell only at a single time point, but we have data across many cells. The deep learning community has recently explored using Schrödinger bridges (SBs) and their extensions in similar settings. However, existing methods either (1) interpolate between just two time points or (2) require a single fixed reference dynamic (often set to Brownian motion within SBs). But learning piecewise from adjacent time points can fail to capture long-term dependencies. And practitioners are typically able to specify a model family for the reference dynamic but not the exact values of the parameters within it. So we propose a new method that (1) learns the unobserved trajectories from sample snapshots across multiple time points and (2) requires specification only of a family of reference dynamics, not a single fixed one. We demonstrate the advantages of our method on simulated and real data. 
\end{abstract}

\section{INTRODUCTION}
\label{sec:intro}
Practitioners are often interested in the possible paths taken by a population of particles moving from one location to another in a given space. For example, biologists want to understand how gene expression, as measured by mRNA levels, changes when normal cells transform into cancer cells. A deeper understanding might aid development of methods to prevent or treat cancer. One can model the dynamics of mRNA concentration in each cell using a stochastic differential equation (SDE), and there is extensive theory on SDEs from trajectories densely sampled in time. However, scientists cannot measure mRNA concentration continuously in time, but rather only at certain time snapshots. Moreover, since measuring mRNA concentration requires destroying the cell, scientists cannot track the trajectory of one cell across multiple times. 

Recent work has demonstrated the potential of using Schr\"odinger bridges (SBs) to infer possible trajectories connecting two time snapshots \citep*{pavon2021data,de2021diffusion, Vargas2021,koshizuka2022neural, Wang2023}. This approach learns a pair of forward-backward SDEs that can transport particles between two time points (forward or backward, respectively). One can then use the learned SDEs to numerically sample from the latent population of trajectories. 
To extend these methods to data with multiple time snapshots, 
one can apply vanilla SBs separately over each pair of consecutive time points.  
But the learned dynamics may fail to capture long-term dependencies, seasonal patterns, and cyclic behaviors.
\citet{lavenant2024toward} show that other natural extensions to multiple time steps \citep{chen2019multi, lavenant2024toward} still decompose into vanilla SBs between adjacent time steps. 

To overcome these issues, \citet{chen2024deep} proposed to instead use the SDEs in SBs to govern particles' velocity, instead of location. As with the methods above, their method allows the dynamics to pass through multiple time snapshots.
In addition, since their method generates smooth trajectories, trajectory information can be shared across time intervals. 

A practical issue with all methods above, including \citet{chen2024deep}, though, is that they require a single pre-defined reference measure. In particular, these methods minimize relative entropy between the process and the reference. 
Ideally, the reference dynamics should be close to the true, latent dynamics. While scientists often have some prior knowledge of how their system works \citep[e.g., the parametric form of the underlying dynamical model, as in][]{pratapa2020benchmarking}, it is rare that they can access all parameter values. Consequently, practitioners often default to using Brownian motion as the reference dynamic. In fact, \citet{schrodinger1932theorie} originally used Brownian motion to model particles moving under thermal fluctuations in a closed system; for a historical review, see \citet{leonard2013survey}. In this context, solving the SB problem is equivalent to solving an entropy-regularized optimal transport problem \citep{cuturi2013sinkhorn}. However, this approach may not be ideal for many open systems with energy intake. For instance, biological systems are such systems and do not necessarily adhere to the same entropy laws as gas molecules under thermal fluctuations. \citet[][Chapter 6]{schrodinger1946life} himself  famously remarked ``What organisms feed upon is \textit{negative} entropy.''

Methods beyond the SB literature face similar (and additional) limitations. 
For instance, TrajectoryNet \citep{tong2020trajectorynet} combines continuous normalizing flows with a soft constraint based on dynamic optimal transport. But it can be interpreted as an SB method with a single reference;
see \citet[sections 3.2 and 4.1]{tong2020trajectorynet}. See \cref{app:related-work} for more related work.

Due to the limitations of existing methods, we propose a new approach that (1) infers the distribution of trajectories from sample snapshots at multiple time points and (2) requires the user to specify only a parametric family of reference dynamics rather than a single one. Our method iterates between two steps. (i) Given our current best guess of the latent dynamics, we learn a piecewise SB and sample the resulting trajectories. (ii) We use the learned SB to refine our best guess of the latent dynamics within the reference family. 
Since the estimate of latent dynamics uses information from all times, we expect our method to share information across time intervals.
In our experiments (\cref{sec:experiments-main}), we find that our method is more accurate than both vanilla SBs and alternatives that nontrivially handle multiple time points. 
Moreover, our method is faster than the latter methods in all experiments.\footnote{Code to reproduce the experiments is available at \href{https://github.com/YunyiShen/SB-Iterative-Reference-Refinement}{https://github.com/YunyiShen/SB-Iterative-Reference-Refinement}.} 

In concurrent work, \citet{zhang2024joint} independently proposes an iterative procedure that allows for a family of reference dynamics. While in our work we use SBs to recover continuous-time evolution of the underlying dynamics, \citet{zhang2024joint} uses an entropic optimal transport framework and aims to recover the underlying dynamics solely at points with observations. And while we allow a general reference family, \citet{zhang2024joint} uses a linear (Ornstein-Uhlenbeck) reference family. We share the same  motivation though: moving away from a fixed reference dynamic to improve trajectory inference.

\section{SETUP AND BACKGROUND}
\label{sec:setup}
We review the inference problem and existing SB work that will be relevant to our method. Though our method is more general, we continue the example above to motivate the setup; that is, suppose we are interested in sampling the trajectories of mRNA concentration in a population of cells.

\textbf{Data.}  We have observations at $\totsteps$ time points. At the $\timeidx$th time, $\timestep{\timeidx}$, we observe data from $\tottrajec_{\timestep{\timeidx}}(> 0)$ cells. The mRNA concentration of the $\sampleidx_{\timeidx}$th cell at $\timestep{\timeidx}$ is $\obsat{\timestep{\timeidx}}^{\sampleidx_{\timeidx}}\in \R^d$. After gathering data for a cell, the cell dies, so each cell is observed only once. More generally, we say that we observe the $(\timeidx, \sampleidx_{\timeidx})$ \emph{particle} just once, at time $\timestep{\timeidx}$. Without loss of generality, we choose the time steps to be unique, increasing, and starting at 0: $0 =\timestep{1} < \timestep{2} < \dots < \timestep{\totsteps} < \infty$. They need not be equally spaced.
We let $\obsatall{\timestep{\timeidx}}$ denote all $\tottrajec_{\timestep{\timeidx}}$ observations at time $\timestep{\timeidx}$. There are $\totalnumparticles = \sum_{\timeidx=1}^\totsteps \tottrajec_{\timestep{\timeidx}}$ total observations.

\textbf{Goal.} 
We assume that, if not measured, each cell's mRNA concentration would have had a continuous trajectory in time. $\responseat{\timet}^{(\timeidx, \sampleidx_{\timeidx}) }$ denotes the trajectory for the $\sampleidx_{\timeidx}$th cell observed at the $\timeidx$th time step. So $\obsat{\timestep{\timeidx}}^{\sampleidx_{\timeidx}} = \responseat{\timet = \timestep{\timeidx}}^{ (\timeidx, \sampleidx_{\timeidx}) }.$ That is, $\obsat{\timestep{\timeidx}}^{\sampleidx_{\timeidx}}$ denotes an evaluation of $\responseat{\timet}^{ (\timeidx, \sampleidx_{\timeidx}) }$ at the observed time $\timestep{\timeidx}$. We assume the trajectories are independent samples from a latent distribution over trajectories; this assumption implies the observations are independent as well. Our goal is to generate samples from the latent distribution over particle trajectories within $\timet \in [0,\timestep{\totsteps}]$, given the observations $\{ \obsat{\timestep{\timeidx}}^{\sampleidx_{\timeidx}} \}_{\timeidx, \sampleidx_{\timeidx}}$.

\textbf{Model.}
We model the latent trajectory of the $(\timeidx, \sampleidx_{\timeidx})$ particle with an SDE driven by a $d$-dimensional Brownian motion $\brownianm^{ (\timeidx,\sampleidx_{\timeidx}) }$, independent across particles:
\begin{equation}
    \de\responseat{\timet}^{(\timeidx, \sampleidx_{\timeidx}) } = \modeldrift(\responseat{\timet}^{ (\timeidx,\sampleidx_{\timeidx}) },t) \de\timet + \sqrt\volatility \de\brownianm^{ (\timeidx,\sampleidx_{\timeidx}) },~~\responseat{\timet = 0}^{ (\timeidx,\sampleidx_{\timeidx}) }\sim \marginal{0}.
    \label{eq:mainsde}
\end{equation}
We assume that the volatility $\volatility$ is known.\footnote{This assumption is common in the SB literature since it ensures that the SB problem is well-posed \citep{Chen2022, Vargas2021, lavenant2024toward}. In our experiments, we use the same fixed volatility value as past SB work, and we find it works well. Estimating volatility from data is an interesting direction for future research.} We assume that the drift $\modeldrift(\cdot,\cdot): \R^{d}\times [0,\timet_{\totsteps}] \to \R^{d}$ and initial marginal distribution $\marginal{0}$ are unknown.

We assume standard SDE regularity conditions. The first assumption below ensures a strong solution to the SDE exists; see \citet[][Chapter 3, Theorem 3.1]{pavliotis2016stochastic}. The second ensures that the process does not exhibit unbounded variability.
\begin{assumption}
\label{assumption-lipschitz}
The drifts are $L$-Lipschitz; i.e., for all $t\in [0,\timet_{\totsteps}]$, $\|\modeldrift(x, t)-\modeldrift(y, t)\| \le L \| x-y \|$, where $\|\cdot \|$ denotes the usual $L^2$ norm of a vector. And we have at most linear growth; i.e., there exist $K<\infty$ and constant $c$ such that $\| \modeldrift(y, t)\| <K \| y \|+c$. 
\end{assumption} 
\begin{assumption}
\label{assumption-bddsecondmoments}
At each time step $\timestep{\timeidx}$, the distribution of the $\tottrajec_{\timestep{\timeidx}}$ particles has bounded second moments.
\end{assumption}

\textbf{Generating sample trajectories.} One approach to generating trajectory samples is to first estimate the unknown drift from data. Using the estimated drift, we can simulate forward and backward in time starting from a particle observation to generate an approximate realization of the SDE solution for $\timet \in [0,\timestep{\totsteps}]$. This approach generates one sample trajectory for each observation, for a total of $\totalnumparticles$ trajectory samples.

More precisely, the drift $\modeldrift$ in \cref{eq:mainsde} defines the forward dynamics of $\responseat{\timet}^{ (\timeidx,\sampleidx_{\timeidx}) }$. Given \cref{eq:mainsde}, the backwards process $\responseat{\timet_{\totsteps}-\timet}^{ (\timeidx,\sampleidx_{\timeidx}) }$ is described by an analogous SDE with the same volatility and a new (backward) drift, $\modeldriftbackward(\cdot,\cdot)$; see \citet{haussmann1986time, follmer2005entropy, cattiaux2023time} for details. Suppose we had access to estimates $\estimatedrift$ and $\estimatebkwdrift$ for the forward and backward drift, respectively. Then given a particle observed at time $\timestep{\timeidx}$, we could simulate a trajectory forward in $\timet \in [\timestep{\timeidx},\timestep{\totsteps}]$ by using \cref{eq:mainsde} with (1) $\modeldrift=\estimatedrift$, (2) the initial distribution equal to a point mass at the observation, and (3) discretized time steps of size $\Delta\timet$. We write the resulting simulated trajectory as $\simulatedresponseat{\timestep{\timeidx}\le \timet\le \timestep{\totsteps}}^{ (i,\sampleidx_{\timeidx}) }=\textrm{\texttt{forwardSDE}}(\estimatedrift, \obsat{\timestep{\timeidx}}^{\sampleidx_{\timeidx}},\Delta \timet)$. 
Similarly, we can sample the part of the trajectory within $\timet \in [0,\timestep{\timeidx}]$ using the backward drift and SDE: $\simulatedresponseat{0\le \timet\le \timestep{\timeidx}}^{(i,\sampleidx_{\timeidx})}=\textrm{\texttt{backwardSDE}}(\estimatebkwdrift, \obsat{\timestep{\timeidx}}^{\sampleidx_{\timeidx}},\Delta \timet)$. In our experiments, we use a standard Euler-Maruyama approach \citep[][Chapter 3.4]{sarkka2019applied}; for full details, see \cref{sec:sampling-routine}.  It remains to estimate the forward and backward drifts.

\textbf{Multi-marginal Schrödinger bridges.}
When we have some prior knowledge about the system's latent dynamics, Schrödinger Bridges \citep[SBs,][]{Wang2023, Vargas2021, lavenant2024toward} can be used to estimate the forward and backward drift. We next review multi-marginal SBs and how their drift estimates arise. The general multi-marginal SB problem \citep{chen2019multi, lavenant2024toward} finds the \emph{bridge} distribution $\bridge$ over trajectories that (1) interpolates between specified marginals while (2) being closest --- in a Kullback--Leibler divergence ($\KL$) sense --- to the prior knowledge as expressed in a \emph{reference} distribution $\refden$. With a slight abuse of notation, we use $\bridge$ and $\refden$ to denote not just the distributions over trajectories but also their respective densities with respect to a Wiener measure. Let the distribution over trajectory points at time $\timestep{\timeidx}$ implied by $\bridge$ be $\popmarginalat{\timestep{\timeidx}}$. Let $\marginal{\timestep{\timeidx}}$ be a desired marginal at time $\timestep{\timeidx}$. Let $[I] := \{1,\ldots, I\}$. Then the multi-marginal SB problem is a constrained optimization problem:\footnote{In the SB literature, this KL divergence is often defined in terms of probability distributions rather than densities. But implicitly, to compute the KL divergence, one uses the densities. So here we directly use the densities.} 
\begin{align}
    \label{eq:SBP-main}
    & \argmin_{\bridge\in \measurefamily: \bridge_0=\refden_0}\KL(\bridge||\refden), \\
    \label{eq:measure-family}
    \textrm{where} & \;\; \measurefamily=\left\{\bridge:\forall i \in [I], \popmarginalat{\timestep{\timeidx}}=\marginal{\timestep{\timeidx}} \right\}
\end{align}
is the family of densities over trajectories satisfying the marginal constraints.

Suppose now that the reference is a distribution over trajectories implied by an SDE as in \cref{eq:mainsde}. Since we assume the volatility $\volatility$ is known, we assume the known volatility is used in the reference. So the reference is specified by its drift $\refdrift$. Under \cref{assumption-lipschitz,assumption-bddsecondmoments}, the resulting probability distribution over trajectories admits a density with respect to the Wiener measure defined by a Brownian motion with volatility $\volatility$ \citep{oksendal2013stochastic, kailath1971structure}. The existence of the density follows from Girsanov's Theorem; see, e.g., Theorem 8.6.3 in \citet{oksendal2013stochastic}) and \cref{app:convex-densities}. We write $\refden = \refden_{\refdrift}$ for the distribution and its density. In this case, \citet{kailath1971structure} shows that not only must the optimal $\bridge$ also solve an SDE of the form in \cref{eq:mainsde}, but it must have the same volatility; $\KL$ will be infinite for other volatilities. Therefore, the optimal $\bridge$ is defined by its drift, and we write $\bridge = \bridge_{\modeldrift}$. Then solving \cref{eq:SBP-main} and \cref{eq:measure-family} is equivalent to finding the best drift for $\bridge_{\modeldrift}$:
\begin{align}
    \label{eq:SBP-drift}
    \argmin_{\modeldrift: \bridge_{\modeldrift} \in \measurefamily,  \bridge_{\modeldrift,0}=\refden_{\refdrift,0} }\KL(\bridge_{\modeldrift}||\refden_{\refdrift} ),
\end{align}
with $\measurefamily$ as in \cref{eq:measure-family}. Practical SB algorithms take the marginals to be the (discrete) empirical distributions of observed data at each time point; that is, $\marginal{\timestep{\timeidx}}$ is constructed from $\obsatall{\timestep{\timeidx}}$, where  $\obsatall{\timestep{\timeidx}}$ represents the collection of all the observations available at time $\timestep{\timeidx}$ \citep{Vargas2021, de2021diffusion}.

Even though \cref{eq:SBP-drift}  appears to leverage information from all time snapshots jointly, \citet{lavenant2024toward, chen2019multi} showed that --- under very mild conditions --- this SB problem is equivalent to a collection of $\totsteps-1$ separate SB problems, each between two adjacent time points. Therefore, practical algorithms for solving this problem return $\totsteps-1$ pairs of forward and backward drifts $\{\estimatedriftstep{i},\estimatebackwarddriftstep{i}\}_{\timeidx=1}^{\totsteps-1}$, where $\estimatedriftstep{i}$ and $\estimatebackwarddriftstep{i}$ are defined over the interval $[\timestep{\timeidx}, \timestep{\timeidx+1}]$. We can construct an estimate $\estimatedrift$ of the forward drift in the time span $\timet \in [0,\timestep{\totsteps}]$ by concatenating these estimates, with an analogous estimate $\estimatebkwdrift$ for the backward drift.

\textbf{Challenges.} Since this approach performs piecewise interpolation between consecutive pairs 
it does not share information across different time intervals. Moreover, this approach requires  practitioners to specify a single reference drift $\refdrift$. Even when scientists know a parametric form for the underlying dynamical system \citep[e.g.,][for mRNA dynamics]{pratapa2020benchmarking}, they rarely have access to all parameter values. So in practice the choice of $\refdrift$ usually corresponds to Brownian motion \citep{Caluya2022, Vargas2021,Chen2022, chen2019multi, chen2024deep}.

\section{OUR METHOD}
\label{sec:method}
Our method allows the specification of a reference family rather than requiring a single reference drift. And it facilitates sharing of information between time intervals. In what follows, we first describe a general optimization setup allowing a reference family. We then establish theoretical guarantees for an iterative approach to solving the optimization. Finally, we provide a practical algorithm that reflects this approach.

\textbf{Optimization with a reference family.} We first generalize the optimization problems from \cref{eq:SBP-main,eq:SBP-drift} to allow a reference family. Let $\modelfamily$ be the chosen reference family of densities over trajectories. Then we replace \cref{eq:SBP-main} with:
\begin{align}
    \label{eq:SB_obj}
    \argmin_{\bridge\in \measurefamily} \min_{\substack{\refden\in\modelfamily:\\ \bridge_{0} = \refden_{0}}}
    \KL(\bridge||\refden),
\end{align}
which reduces to \cref{eq:SBP-main} when $\modelfamily$ has a single element.
Next we restrict the elements of the reference family to be distributions over trajectories implied by the SDE in \cref{eq:mainsde}, with shared volatility $\volatility$. As before, it must be the case that the minimizer of \cref{eq:SB_obj} also solves an SDE of the form in \cref{eq:mainsde}, with the same volatility $\volatility$. So we can write $\refden=\refden_{\refdrift}$ and $\bridge=\bridge_{\modeldrift}$. Then solving \cref{eq:SB_obj} is equivalent to solving for the optimal drift:
\begin{align}
    \label{eq:SBP_obj_drift}
    \argmin_{\modeldrift: \bridge_{\modeldrift} \in \measurefamily} \min_{\substack{\refdrift:\refden_{\refdrift}\in\modelfamily,\\\bridge_{\modeldrift,0}=\refden_{\refdrift,0}}} \KL(\bridge_{\modeldrift}||\refden_{\refdrift} )
\end{align}

\textbf{An iterative approach.} To solve the optimization problem in \cref{eq:SBP_obj_drift}, we propose an iterative approach. Namely, we propose to iterate between the following two steps after initializing with some $\refdrift^{(0)}$ and $\iteralgo=1$. 
\begin{align}
    \drift^{(\iteralgo)} &= \argmin_{ \drift: \bridge_{\drift} \in \measurefamily} \, \KL(\bridge_{\drift} \| \refden_{\refdrift^{(\iteralgo-1)}})\label{eq:first-proj}\\
    \refdrift^{(\iteralgo)} &= \argmin_{\refdrift: \refden_{\refdrift}\in\modelfamily} \, \KL(\bridge_{\drift^{(\iteralgo)}} \| \refden_{\refdrift})\label{eq:second-proj}
\end{align}

By construction, this iterative scheme will monotonically decrease the objective in \cref{eq:SBP_obj_drift} at each step. Our next result shows that, under additional assumptions, the scheme converges to the optimum.
\begin{restatable}{myProposition}{iterkl}
\label{prop:iter-kl}
Suppose $\modelfamily$ is a convex set of densities over trajectories implied by the SDE in \cref{eq:mainsde}; suppose all densities have shared volatility $\volatility$ and the same marginal distribution at time $0$.
Suppose that all SDEs satisfy~\cref{assumption-lipschitz,assumption-bddsecondmoments}.
Suppose $\exists C < \infty$ such that, for every drift $\drift$ with $\bridge_{\modeldrift} \in \measurefamily$ or $\refden_{\modeldrift}\in\modelfamily$,
\begin{align*}
&\sup_{(x,t) \in \mathbb{R}^n \times [0,\timestep{\totsteps}]} \| \drift(x, t)\|_{\infty} \leq C 
\end{align*}
Take any initialization $\refdrift^{(0)}$. If $\drift^{(\iteralgo)}$ and $\refdrift^{(\iteralgo)}$ are computed by recursively applying \cref{eq:first-proj,eq:second-proj}, we have
\[
    \lim_{\iteralgo \to \infty} \KL(\bridge_{\drift^{(\iteralgo)}} \| \refden_{\refdrift^{(\iteralgo)}}) = \inf_{\substack{\bridge_{\drift} \in \measurefamily \\ \refden_{\refdrift} \in \modelfamily}} \KL(\bridge_{\drift} \| \refden_{\refdrift})
\]
\end{restatable}
We prove this result in \cref{app:iter_proj_theory} by adapting standard arguments from the theory of iterative projections \citep{csiszar1975divergence, csiszar2004information, benamou2015iterative}. Along the way, we prove and use that $\measurefamily$ is convex (\cref{lemma:convexD}). We do not necessarily expect the reference family $\modelfamily$ to be convex though. In practice, then, our result guarantees convergence to a local, rather than global, minimum of the KL. We also note that convergence of the KL need not imply convergence of the arguments of the KL; see \cref{sec:identifiability} for further discussion of potential identifiability issues.

\textbf{A practical algorithm.}
We cannot solve \cref{eq:first-proj,eq:second-proj} exactly. We next detail how we derive approximate solutions, $\estimatedriftstep{{(\iteralgo)}}$ and $\estimaterefdriftstep{{(\iteralgo)}}$. We summarize our full method in Algorithm~\ref{algo:sb-irr-algo}.

\SetKwComment{Comment}{/* }{ */}
\begin{algorithm}[!t]
    \caption{Our method iterates between (1) estimating the forward and backward bridge drifts given the current best reference guess and (2) using simulated trajectories from the current bridge drift estimates to find a new best reference guess.}
    \label{algo:sb-irr-algo}
    \KwIn{ 
    $\{\obsatall{\timestep{\timeidx}}\}_{\timeidx=1}^{\totsteps}$,
    iteration count $K$, $\Delta \timet$}
    
    $\estimaterefdriftstep{(0)} \gets 0$, $k\gets 0$\ \\
    \While{ $k<K$}{
        $\iteralgo = \iteralgo+1$ \\
        \For{$i=1,\dots,\totsteps-1$}{
            $\estimatedriftstep{\timeidx(\iteralgo)},\estimatebackwarddriftstep{\timeidx(\iteralgo)}=\textrm{\texttt{Forward-Backward-SB}}(\obsatall{\timestep{\timeidx}}, \obsatall{\timestep{\timeidx+1}}||\estimaterefdriftstep{(\iteralgo-1)})$\ 
        }
        $\textrm{SampleTrajectories}=\{\}$\ 
        
        \For{$\timeidx=1,\dots,\totsteps-1$}{
        \For{$\sampleidx_{\timeidx}=1,\dots,\tottrajec_{\timestep{\timeidx}}$}{
            
            $\simulatedresponseat{0\le \timet\le \timestep{\timeidx}}^{(i,\sampleidx_{\timeidx})}\!= \textrm{\texttt{backwardSDE}}(\estimatebackwarddriftstep{(\iteralgo)}, \obsat{\timestep{\timeidx}}^{\sampleidx_{\timeidx}},\Delta \timet)$ 
            
            $\simulatedresponseat{\timestep{\timeidx}< \timet\le \timestep{\totsteps}}^{ (i,\sampleidx_{\timeidx}) }=\textrm{\texttt{forwardSDE}}(\estimatedriftstep{(\iteralgo)}, \obsat{\timestep{\timeidx}}^{\sampleidx_{\timeidx}},\Delta \timet)$
            
            SampleTrajectories.append($\simulatedresponseat{0 \le \timet\le \timestep{\totsteps}}^{(i,\sampleidx_{\timeidx}) }$)
            
        }
        }
        $\estimaterefdriftstep{{(\iteralgo)}} \gets$ \textrm{\texttt{MLEfit}}(SampleTrajectories)
    }
    \KwOut{$\estimatedriftstep{(K)}, \estimatebackwarddriftstep{(K)}$ to generate trajectories}
\end{algorithm}

\textbf{Algorithm step 1.} First, consider \cref{eq:first-proj}, where we optimize the bridge model given the current reference. This step is a multi-marginal SB problem, with reference dynamic $\refden_{\estimaterefdriftstep{{(\iteralgo)}}}$. As discussed above, the problem reduces to standard SB problems pairwise between time points. There are many implementations that approximately solve the SB problem: e.g., a regression method \citep{Vargas2021}, score matching \citep{wang2021deep}, or a proximal method \citep{Caluya2022}. We denote an implementation's output between time points $\timestep{\timeidx}$ and $\timestep{\timeidx+1}$ as $\estimatedriftstep{\timeidx(\iteralgo)},\estimatebackwarddriftstep{\timeidx(\iteralgo)}=\textrm{\texttt{Forward-Backward-SB}}(\obsatall{\timestep{\timeidx}}, \obsatall{\timestep{\timeidx+1}}||\refdrift^{(\iteralgo-1)})$. 
In our code implementation, we follow the regression-based method of \citet{Vargas2021}; see \cref{sec:first_projection_implementation}.
However, our framework is in principle compatible with any solver that (i) supports a general reference measure beyond Brownian motion and (ii) outputs forward and backward drifts for the particle evolution. For a more comprehensive discussion of alternative methods for this step, we refer the reader to \cref{app:sb-methods-discussion}.

\textbf{Algorithm step 2.} The second problem, \cref{eq:second-proj}, does not arise in standard SBs. Our proposed methodology for an approximate solution has two parts, which we detail below: (i) we sample trajectories given the forward and backward bridge drift estimates from the first step, and (ii) we solve a maximum likelihood problem given these sample trajectories.

\begin{figure*}[!ht]
    \centering
    \includegraphics[width =.95\textwidth]{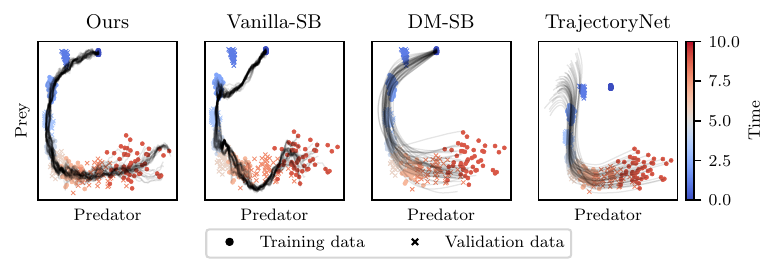}
    \caption{Comparison on the Lotka-Volterra synthetic data (\cref{sec:LVmain}) with 5 training times, 4 validation times, and 50 observations per time. Each plot shows 50 simulated trajectories, originating from particles at one time end point (three left plots: first time; right plot: final time); see ``Measuring Accuracy'' in \cref{sec:experiments-main}.  
    }
    \label{fig:LV-main}
\end{figure*}

\textbf{Algorithm step 2.i.} From step 1, we have forward and backward bridge drift estimates $\estimatedriftstep{(\iteralgo)},\estimatebackwarddriftstep{(\iteralgo)}$. Recall ``Generating sample trajectories'' within \cref{sec:setup}; for each observation $\obsat{\timestep{\timeidx}}^{\sampleidx_{\timeidx}}$, we can simulate a trajectory in $\timet \in [0,\timestep{\totsteps}]$ via $\simulatedresponseat{0\le \timet\le \timestep{\timeidx}}^{(i,\sampleidx_{\timeidx})}= \textrm{\texttt{backwardSDE}}(\estimatebackwarddriftstep{(\iteralgo)}, \obsat{\timestep{\timeidx}}^{\sampleidx_{\timeidx}},\Delta \timet)$ and $\simulatedresponseat{\timestep{\timeidx}< \timet\le \timestep{\totsteps}}^{ (i,\sampleidx_{\timeidx}) }=\textrm{\texttt{forwardSDE}}(\estimatedriftstep{(\iteralgo)}, \obsat{\timestep{\timeidx}}^{\sampleidx_{\timeidx}},\Delta \timet)$. Then we have $\totalnumparticles$ simulated trajectories.

\textbf{Algorithm step 2.ii.}
By expanding $\KL$ and dropping terms where $\refden_{\refdrift}$ does not appear, we can rewrite \cref{eq:second-proj} as follows; see \cref{app:inverseproj} for full details and derivation of this step of the algorithm. 
\begin{align}
    \refdrift^{(\iteralgo)} &= \argmin_{\refdrift: \refden_{\refdrift}\in\modelfamily} \, \KL(\bridge_{\drift^{(\iteralgo)}} \| \refden_{\refdrift}) \\
    &= \argmax_{\refdrift: \refden_{\refdrift}\in\modelfamily} \mathbb{E}_{\bridge_{\drift^{(\iteralgo)}}} \log \refden_{\refdrift}
    \label{eq:max_likelihood}
\end{align}
We can approximate the exact objective in \cref{eq:max_likelihood}
by replacing the expectation $\mathbb{E}_{\bridge_{\drift^{(\iteralgo)}}}$ with an empirical average over the trajectories $\{\simulatedresponseat{0\le \timet\le \timestep{\totsteps}}^{(\timeidx,\sampleidx_{\timeidx})}\}_{\timeidx,\sampleidx_{\timeidx}}$, which are simulated from $\bridge_{\drift^{(\iteralgo)}}$. The resulting optimization problem over this approximate objective can be seen as maximizing the (log) likelihood of the simulated trajectories $\{\simulatedresponseat{0\le \timet\le \timestep{\totsteps}}^{(\timeidx,\sampleidx_{\timeidx})}\}_{\timeidx,\sampleidx_{\timeidx}}$ over the choice of $\refdrift$ in $\refden_{\refdrift}$.
Since the simulated trajectories are obtained by a standard Euler-Maruyama approach with sampling rate $\Delta t$, they can be represented by a Gaussian autoregressive process.
We show in \cref{app:inverseproj} that the optimization problem therefore reduces to a least squares scheme, where the square errors arise from the Gaussian log likelihoods. Given the connection to maximum likelihood estimation (MLE), we let \texttt{MLEfit} denote the function that takes in sample trajectories and outputs our resulting estimate $\estimaterefdriftstep{(\iteralgo)}$ of the reference drift $\refdrift^{(\iteralgo)}$.

\section{EXPERIMENTS}
\label{sec:experiments-main}
In simulated and real experiments, we find that our method gives more accurate predictions than alternatives and is substantially faster than existing methods that share information across time intervals. Each of our experiments explores a different application with a different natural reference family.

\textbf{Baselines.} We compare to three other methods, all described in \cref{sec:intro}. (1) A vanilla multimarginal SB (\textit{\vanilla{}}) with a Brownian motion reference. See \cref{sec:vanilla-app} for implementation details.
(2) The Deep Momentum multimarginal Schr\"odinger bridge (\textit{\dmsbname{}}) \citep{chen2024deep}. We use the authors' code at \href{https://github.com/TianrongChen/DMSB.}{\nolinkurl{https://github.com/TianrongChen/DMSB}}; see \cref{sec:dmsb-app} for details. (3) \textit{TrajectoryNet} \citep{tong2020trajectorynet}. We use the code at \href{https://github.com/KrishnaswamyLab/TrajectoryNet}{\nolinkurl{https://github.com/KrishnaswamyLab/TrajectoryNet}}; see \cref{app:trajnetdetails}. In all SB methods, we set the volatility $\volatility$ to 0.1, as suggested by \citet{Vargas2021}. See \cref{app:train-hyperparams-ref} for implementation details of our method. 

\textbf{Measuring accuracy.} In each experiment, we start with a collection of data at an odd total number of time points. We train each method using data from odd-indexed times (including the first and last time points); we use even-indexed times as held-out validation data. At a high level, we measure accuracy by how well the empirical distribution of a method's simulated trajectories corresponds to held-out data at each validation time point. We evaluate the performance both visually and via numerical summaries of error. 

Since all methods estimate drift, it should be possible to simulate trajectories from all training particles for all methods. However, the \dmsbname{} code provides trajectories originating only from the first-time-point particles, and the TrajectoryNet code provides trajectories originating only from the final-time-point particles. Therefore, to provide a more direct comparison of trajectory quality, we plot only the subset of trajectories provided by our method and \vanilla{} that arise from the first-time-point particles.

We report numerical summaries for our method both for these restricted trajectories and also for trajectories using all particles. In practice, we recommend practitioners use our full method with all particles; we consider restricted trajectories only for comparison to existing, restricted code.
While we focus on the Earth Mover's Distance (EMD) in the main text, we also compute the Maximum Mean Discrepancy (MMD) in \cref{app:further-experiments}. We discuss these choices in \cref{app:metrics}.

In each table entry, we report the mean and standard deviation of each result across 10 random seeds.
We report results on the restricted set of trajectories in the first rows of each table, and results over all possible trajectories in the final rows.
We highlight (in green) the best restricted result (across methods), and any other (restricted) method whose one-standard-deviation confidence interval
overlaps the mean of the best (restricted) method. We highlight (in light blue) any all-trajectory method that beats the best restricted result or any all-trajectory method whose one-standard-deviation confidence interval
overlaps the mean of the best restricted method. We provide more details in \cref{app:tiebreaking}. Our conclusions generally agree across all metrics.

\begin{figure*}[!ht]
    \centering
    \includegraphics[width = .95\textwidth]{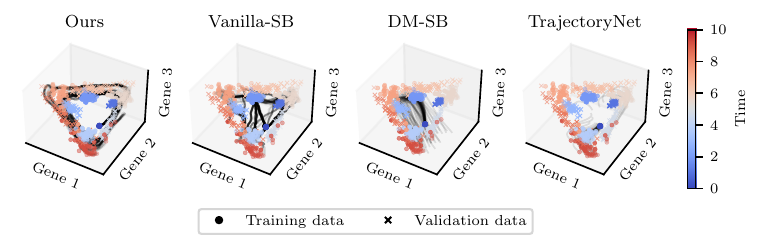}
    \caption{\label{fig:repr-main} 
    Comparison on the repressilator synthetic data (\cref{sec:represmain}) with 6 training times, 5 validation times, and 50 observations per time. Each plot shows 50 simulated trajectories, originating from particles at one time end point (three left plots: first time; right plot: final time).}
\end{figure*}

\textbf{Measuring runtime.} We describe our run time comparisons in detail, and provide full running time results, in \cref{app:timing-table}. We emphasize that implementation details affect runtime, which can be hard to interpret for iterative SB methods without clear stopping criteria. For instance, our method may slow down with more iterations and samples due to our piece-wise SB solver choice, but could also be accelerated with faster piece-wise SB solvers. Similarly, other methods may benefit from fewer iterations or improved SDE solvers. Therefore, these reported runtimes should be interpreted in the context of each experiment's specific setup.

\subsection{Lotka-Volterra} 
\label{sec:LVmain}
We generate synthetic data from a stochastic Lotka-Volterra predator-prey model; we use the same parametric dynamical system as the reference family in our method. We take 5 training and 4 validation time points, with 50 observations per time point. See \cref{app:lv-app} for more details.

\textbf{Accuracy.} In \cref{fig:LV-main}, we see that, since \vanilla{} learns drift by interpolating between each pair of time points, it misses the curvature of the dynamics. TrajectoryNet misses the curvature of the initial time points. While both \dmsbname{} and our method capture the overall curvature, the enforced smoothness of \dmsbname{}'s trajectories lead to substantial mass away from the validation data (as in the lower left corner) and worse EMD performance; see \cref{tab:LV50_EMD_appendix}.

\textbf{Runtime.} Across the 10 seeds, we find the following runtimes in hours, reported as mean $\pm$ one standard deviation: \dmsbname{} 7.62$\pm$3.18, TrajectoryNet 10.96$\pm$0.81, ours 0.61$\pm$0.12, and \vanilla{} 0.06$\pm$0.01. Of the methods incorporating information across multiple time intervals, our method is (on average across runs) the fastest.
\vanilla{} is the fastest due to its relative simplicity. Since \vanilla{} is a subroutine in our method, we expect our method to be about $K$ times as expensive. TrajectoryNet is built on constrained continuous normalizing flows, which are known to be computationally intensive \citep{grathwohl2018ffjord}.
We conjecture that \dmsbname{} faces computational challenges due to learning dynamics in a larger space (velocity and location vs.\ just location), with no direct observations in the velocity part of the space. In the context of generative modeling using diffusion, \citet{dockhorn2021score} also augment their model with particle velocities and face increased training time.

\subsection{Repressilator} 
\label{sec:represmain}
We generate synthetic data based on a model that captures the circadian rhythm in cyanobacteria \citep{nakajima2005reconstitution}. Three coupled SDEs model mRNA levels of three genes that cyclically suppress each other's synthesis. We use the same parametric dynamics as the reference family in our method. We take 6 training and 5 validation times, with 50 observations per time. See \cref{app:repr-app} for more details.

\textbf{Accuracy.} In \cref{fig:repr-main}, we see that \vanilla{}, \dmsbname{}, and TrajectoryNet all fail to capture the cyclic nature of these dynamics. But our method accurately captures the curvature. The EMD summaries agree that our method outperforms the alternatives; see \cref{tab:repres50_EMD_appendix}.

\textbf{Runtime.} Across the 10 seeds, we find the following runtimes in hours (mean $\pm$ one standard deviation): \dmsbname{} 15.63$\pm$0.12, TrajectoryNet 9.86$\pm$0.43, ours 2.43$\pm$0.60, and \vanilla{} 0.23$\pm$0.05. Our method is (on average across runs) faster than all the baselines sharing information across intervals.
Again, our method offers substantial accuracy gains at a lower computational cost.

\begin{figure*}[!ht]
    \centering
    \includegraphics[width =.97\textwidth]{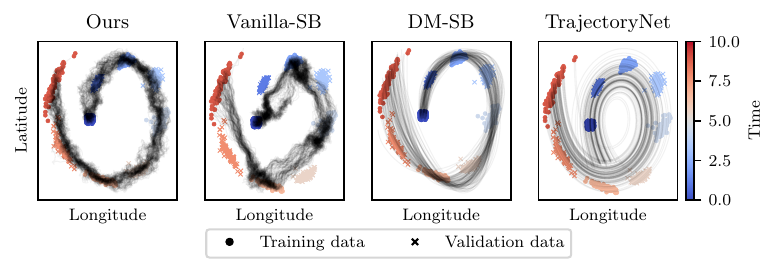}
    \caption{    Comparison on the Gulf of Mexico data (\cref{sec:gommain}) with 5 training times, 4 validation times, and approximately 111 observations per time. Each plot shows approximately 111 simulated trajectories, originating from particles at one time end point (three left plots: first time; right plot: final time). 
    }
    \label{fig:GoMsmall-main}
\end{figure*}

\subsection{Ocean currents in the Gulf of Mexico}
\label{sec:gommain}
In our remaining two analyses, our reference family is mis-specified; that is, since we use real data, the reference family cannot exactly match the data-generating process. 
We first use real ocean-current data from the Gulf of Mexico: we use high-resolution (1 km) bathymetry data from a HYbrid Coordinate Ocean Model (HYCOM) reanalysis\footnote{Dataset available at \url{https://www.hycom.org/data/gomb0pt01/gom-reanalysis}.} to obtain a velocity field around what appears to be a vortex feature. We simulate particles (representing buoys or ocean debris) that evolve according to this field. We observe a total of 1000 particles across 5 training and 4 validation times. In our method, we use a reference parametric family representing a vortex with unknown center,
direction, and shape. See \cref{app:GoM-app-reference} for full details.  

\textbf{Accuracy.} In \cref{fig:GoMsmall-main}, \vanilla{} produces polygon-like ``kinks" --- i.e., sharp turns in trajectories --- at training time points. These kinks represent a failure to capture a smooth vortex motion aligned with fluid dynamics principles. These artifacts arise because \vanilla{}'s drift is learned strictly from pairwise time intervals, instead of incorporating information from the entire time horizon. \dmsbname{} and TrajectoryNet generate smooth trajectories that are notably far from the data at the final validation time point. Our method's trajectories track the validation data closely. Our EMD results in \cref{tab:GoMsmall_EMD_appendix} reflect that both our method and \dmsbname{} are close to the data at the first 3 validation time points. But \dmsbname{} is about twice as far (in EMD) relative to our method (or to either method at other validation points) at the final validation point. We provide results for an additional experiment on particles farther from the center of the vortex in \cref{app:GoM-app-results}.

\textbf{Runtime.} Across the 10 seeds, we find the following runtimes in hours (mean $\pm$ one standard deviation): \dmsbname{} 15.44$\pm$0.02, TrajectoryNet 7.44$\pm$0.25, ours 4.67$\pm$0.66, and \vanilla{} 0.43$\pm$0.01. Of the methods sharing information across multiple time intervals, our method is (on average across runs) the fastest.

\subsection{Single-cell sequencing} 
\label{sec:scrnamain}
We next consider two single-cell sequencing datasets. We follow  \citet{tong2020trajectorynet,chen2024deep} in analyzing data from \citet{moon2019visualizing} on embryoid body (EB) cells. We analyze an additional dataset from \citet{chu2016single} on human embryonic stem cells (hESC). Both datasets promise insight into the dynamic process of stem cell differentiation. 
We use the same pre-processing pipeline of \citet{tong2020trajectorynet} for the EB data, and we apply this pipeline to the hESC data.
The EB data has 3 training and 2 validation time points. We subsample it to have 300 cells at each time.
The hESC dataset initially has 6 time points. We remove the final time point so that we start and end with training times. The resulting training times have 92, 66, 138 cells, and the validation times have 102, 172 cells, respectively.

For the reference family input to our method, we use a gradient field family, following \citet{wang2011quantifying, weinreb2018fundamental, lavenant2024toward}. This family is motivated by Waddington's famous analogy between cellular differentiation and a marble rolling down a potential surface \citep{waddington2014strategy}. 
To parameterize the gradient field in our experiment, we use a multilayer perceptron (MLP), an architecture commonly used for gradient field representations \citep[e.g.,][]{greydanus2019hamiltonian,lin2023computing}. Our MLP consists of three hidden layers with sizes 128, 64, and 64, interconnected by ReLU activation functions. We choose this specific configuration, including sizes and training hyperparameters, by doing a grid search detailed in \cref{app:eb_implementation}. 

\textbf{Accuracy.} We see from \cref{tab:EB_EMD} that, when comparing the quality of only those trajectories generated from a single end point in time, \dmsbname{} outperforms the alternatives, including our method. However, when allowed to generate trajectories from all particles, \vanilla{} and our method outperform the single-time trajectory options (as expected) and perform comparably to each other. Indeed, when we look at the data itself (\cref{fig:eb-app} for EB, \cref{fig:hesc-app} for hESC), it appears that there is not much pattern in the data on the time scale at which the data is sampled; the EB data in particular appears to move in a single direction. In neither the EB nor hESC case do we see a clear and meaningful visual pattern that any method picks up. We emphasize that the number of training time points is only three, so there is not much pattern that could be picked up. With all of these considerations in mind, we conjecture that \dmsbname{} and TrajectoryNet would perform about as well as the \vanilla{} method or our method in these two cases if they were used to generate trajectories from all training particles; we are not able to confirm with the existing code.

\begin{table*}[!ht]
    \centering
    \begin{tabular}{lcccc}
    \hline
       & \multicolumn{2}{c}{EB} & \multicolumn{2}{c}{hESC} \\
        Method & EMD $t_2$ & EMD $t_4$ & EMD $t_2$ & EMD $t_4$ \\
    \hline
        \vanillaonetime & 1.49 $\pm$ 0.063 & 1.55 $\pm$ 0.034 & 1.47 $\pm$ 0.088 & 1.97 $\pm$ 0.169\\
         \dmsb & \cellcolor{green!25}\textbf{1.13 $\pm$ 0.082} & \cellcolor{green!25}\textbf{1.45 $\pm$ 0.16} & \cellcolor{green!25}\textbf{1.10 $\pm$ 0.066} & 1.51 $\pm$ 0.11\\
         \trajnet & 2.03 $\pm$ 0.04 & 1.93 $\pm$ 0.08 & 1.30 $\pm$ 0.04 & 1.93 $\pm$ 0.05 \\
         \oursonetime & 1.27 $\pm$ 0.028 & 1.57 $\pm$ 0.048 & \cellcolor{green!25}\textbf{1.08 $\pm$ 0.12} & \cellcolor{green!25}\textbf{1.33 $\pm$ 0.084} \\
    \hline
        \vanillaalltime & \cellcolor{blue!10}\textbf{1.12 $\pm$ 0.031} & \cellcolor{blue!10}\textbf{1.12 $\pm$ 0.023} & \cellcolor{blue!10}\textbf{0.72 $\pm$ 0.017} & \cellcolor{blue!10}\textbf{1.27 $\pm$ 0.043} \\
        \oursalltime & \cellcolor{blue!10}\textbf{0.96 $\pm$ 0.019} & \cellcolor{blue!10}\textbf{1.19 $\pm$ 0.017} & \cellcolor{blue!10}\textbf{0.71 $\pm$ 0.031} & \cellcolor{blue!10}\textbf{1.25 $\pm$ 0.076} \\
    \hline
    \end{tabular}
    \caption{Earth mover's distance (mean $\pm$ standard deviation) at 2 validation times for the EB and hESC data. The first four rows use only trajectories generated from one time endpoint. The final two rows use all trajectories. }
    \label{tab:EB_EMD}
\end{table*}

\textbf{Runtime.} First consider EB. Across the 10 seeds, we find the following runtimes in hours (mean $\pm$ one standard deviation): \dmsbname{} 15.54$\pm$0.41, TrajectoryNet 10.19$\pm$0.37, ours 0.38$\pm$0.05, and \vanilla{} 0.03, with a standard deviation less than 0.01. Of the methods sharing information across multiple time intervals, our method is (on average across runs) the fastest.
Next consider hESC. We find the following runtimes in hours: \dmsbname{} 15.40$\pm$0.08, TrajectoryNet 8.00$\pm$0.49, ours 0.56$\pm$0.04, and \vanilla{} 0.05, with a standard deviation less than 0.01. Also for this experiment, our method is (on average across runs) the fastest.

\section{DISCUSSION}
\label{sec:discussion}
We demonstrated that our proposed iterative method effectively reconstructs population trajectories from sample snapshots at multiple time points, achieving greater accuracy and lower computational cost compared to existing approaches. A key advantage of our method is the reuse of a single refined reference across all intervals, allowing temporal information to propagate from one drift update to the next. However, when the data contain too few time points or exhibit weak temporal structure, one should not expect substantial improvements over pairwise methods --- though our approach should not perform worse in these cases.

Several interesting directions remain for future work. First, determining the precise conditions under which a latent SDE is identifiable from marginal samples remains an open problem; see \cref{sec:identifiability} for further discussion. Second, while our current method treats the volatility $\volatility$ as fixed and constant across each iteration, one could introduce an outer loop that iteratively re-estimates $\volatility$ from the data. In this approach, each iteration of our algorithm would still use a fixed $\volatility$, preserving the standard KL-based formulation. Afterward, one would update $\volatility$ --- for instance, by matching the residual variance of the SB-imputed trajectories --- before the next round of inference. However, it is unclear how this extra loop would affect overall convergence, since each update to $\volatility$ changes the underlying SB optimization problem. This is an interesting direction for future work.

Additionally, our iterative refinement procedure may converge to suboptimal local minima. In practice, incorporating multiple random initializations or performing sensitivity analyses could help assess the robustness of the solution, and we plan to explore these aspects in future work. Our method could also be combined with existing approaches (including \dmsbname{} or TrajectoryNet) to enforce constraints such as trajectory smoothness. While our current formulation assumes no observational noise, we may incorporate noise using ideas from \citet{Wang2023}. Finally, we hope to extend our method to make forecasts, allowing predictions beyond the observed time points.

\acknowledgments{The authors are grateful to Gabriele Corso, Vishwak Srinivasan, and Théo Uscidda for useful discussions and comments. This work was supported in part by the Office of Naval Research under
grant N00014-20-1-2023 (MURI ML-SCOPE) and by the NSF TRIPODS program (award DMS-2022448).}

\bibliography{references.bib}

\clearpage

\newboolean{accepted}
\setboolean{accepted}{false}  

\ifthenelse{\boolean{accepted}}{}{
\section*{Checklist}

The checklist follows the references. For each question, choose your answer from the three possible options: Yes, No, Not Applicable.  You are encouraged to include a justification to your answer, either by referencing the appropriate section of your paper or providing a brief inline description (1-2 sentences). 
Please do not modify the questions.  Note that the Checklist section does not count towards the page limit. Not including the checklist in the first submission won't result in desk rejection, although in such case we will ask you to upload it during the author response period and include it in camera ready (if accepted).

 \begin{enumerate}

 \item For all models and algorithms presented, check if you include:
 \begin{enumerate}
   \item A clear description of the mathematical setting, assumptions, algorithm, and/or model. [\textbf{Yes} -- See \cref{sec:setup} and \cref{sec:method}.]
   \item An analysis of the properties and complexity (time, space, sample size) of any algorithm. [\textbf{Yes} -- We discuss our algorithm and baselines runtime for each experiment in the ``Runtime" subparagraphs in \cref{sec:experiments-main}.]
   \item (Optional) Anonymized source code, with specification of all dependencies, including external libraries. [\textbf{Yes} -- We provide a link to access our code at the start of \cref{app:train-hyperparams-ref}.]
 \end{enumerate}

 \item For any theoretical claim, check if you include:
 \begin{enumerate}
   \item Statements of the full set of assumptions of all theoretical results. [\textbf{Yes} -- We state or reference all necessary assumptions in the respective result statement.]
   \item Complete proofs of all theoretical results. [\textbf{Yes} -- See \cref{app:iter_proj_theory}.]
   \item Clear explanations of any assumptions. [\textbf{Yes} -- For each assumption through the paper we explain why it is needed and why it is a reasonable assumption to make.]     
 \end{enumerate}

 \item For all figures and tables that present empirical results, check if you include:
 \begin{enumerate}
   \item The code, data, and instructions needed to reproduce the main experimental results (either in the supplemental material or as a URL). [\textbf{Yes} -- We provide a link to access our code, data and instructions needed to reproduce the main experimental results in \cref{app:train-hyperparams-ref}.]
   \item All the training details (e.g., data splits, hyperparameters, how they were chosen). [\textbf{Yes} -- We discuss these experimental details for each experiment in \cref{app:further-experiments}.]
         \item A clear definition of the specific measure or statistics and error bars (e.g., with respect to the random seed after running experiments multiple times). [\textbf{Yes} -- We discuss metrics in \cref{app:metrics} and error bars in \cref{app:tiebreaking}.]
         \item A description of the computing infrastructure used. (e.g., type of GPUs, internal cluster, or cloud provider). [\textbf{Yes} -- We discuss computing infrastructure at the start of \cref{app:timing-table}.]
 \end{enumerate}

 \item If you are using existing assets (e.g., code, data, models) or curating/releasing new assets, check if you include:
 \begin{enumerate}
   \item Citations of the creator If your work uses existing assets. [\textbf{Yes}]
   \item The license information of the assets, if applicable. [\textbf{Yes} -- We provide license information in the appendices where we discuss the detail of the baselines code (\cref{sec:dmsb-app} and \cref{app:trajnetdetails}) and dataset used in each experiment with real data (\cref{app:GoM-app-exp-details}, \cref{app:singlecellsetup}).]
   \item New assets either in the supplemental material or as a URL, if applicable. [\textbf{Yes} -- We provide a link to our code in \cref{app:train-hyperparams-ref}.]
   \item Information about consent from data providers/curators. [\textbf{Not Applicable}]
   \item Discussion of sensible content if applicable, e.g., personally identifiable information or offensive content. [\textbf{Not Applicable}]
 \end{enumerate}

 \item If you used crowdsourcing or conducted research with human subjects, check if you include:
 \begin{enumerate}
   \item The full text of instructions given to participants and screenshots. [\textbf{Not Applicable}]
   \item Descriptions of potential participant risks, with links to Institutional Review Board (IRB) approvals if applicable. [\textbf{Not Applicable}]
   \item The estimated hourly wage paid to participants and the total amount spent on participant compensation. [\textbf{Not Applicable}]
 \end{enumerate}

 \end{enumerate}
}

\newpage
\onecolumn
\section*{Supplemental Material}
\appendix
\section{Additional Related Work}
\label{app:related-work}
In this section, we expand our related work discussion beyond the Schrödinger bridges literature to works that are closely related to our problem.

\textbf{Generative modeling.} Recently, there has been significant progress in the literature on modeling flows or transports between probability distributions. These approaches are built on a variety of frameworks, such as score-based generative modeling \citep{song2019generative, song2020score}, diffusion models \citep{ho2020denoising}, Schrödinger bridges \citep{de2021diffusion,pavon2021data,Vargas2021,Wang2023}, continuous normalizing flows \citep{chen2018neural,grathwohl2018ffjord}, and flow matching \citep{lipman2022flow}. However, the majority of these works focus on generative modeling, where the goal is to transform a noise distribution, such as a Gaussian, into a data distribution to generate samples. Moreover, they typically address transport between distributions at only two time points within one being typically Gaussian. In contrast, our focus is not on learning how to generate data from noise, but on learning and reconstructing trajectories that capture the underlying system dynamics as they evolve across multiple time steps. 

\textbf{Trajectory inference.} Recent studies have been focused on interpolating the trajectories of individual samples at multiple time points with optimal  transport \citep{schiebinger2019optimal, yang2018scalable}. However in certain situations just mere interpolation is not enough, since the learned dynamics may fail to capture long-term dependencies, seasonal patterns, and cyclic behaviors. Other works in the optimal transport literature helped to overcome this issue: \citet{hashimoto2016learning} proposed to reconstruct trajectories using a particular regularized recurrent neural network (RNN) with Wasserstein gradient flow loss. \citet{bunne2022proximal} expanded this approach by modeling the evolution of the trajectories of interest using the Jordan-Kinderlehrer-Otto (JKO) scheme. In particular, the authors developed \textit{JKOnet}, a neural architecture that can compute the JKO flow, from which the energy landscape that governs the population's dynamics can be learnt. \textit{TrajectoryNet} by \citet{tong2020trajectorynet} combines ideas from dynamic optimal transport with continuous normalizing flows (CNFs). In particular, the authors use a CNF to generate continuous-time non-linear sample trajectories from multiple time points. And they also propose a regularization that encourages a straight trajectory based on dynamic optimal transport theory. \citet{huguet2022manifold} recently proposed another flow-based method,
Manifold Interpolation Optimal-transport Flow (\textit{MIOFlow}). The main idea of this work is to solve the flow problem in a manifold embedding. The authors use a neural ODE \citep{chen2018neural} to transport a sampling of high dimensional data points between time points such that (1) the transport occurs on a manifold defined by samples, (2) the transport is penalized
to agree with measured time points using Wasserstein, and (3) the transport is stochastic.

All of these methods can be viewed as optimizing an entropy-regularized (unbalanced) optimal transport loss between observed and simulated samples, requiring backpropagation through optimal transport solvers \citep{cuturi2013sinkhorn, cuturi2022optimal}. Essentially, these methods solve a Schrödinger bridge problem, each utilizing a specific reference measure based on the chosen method. For example, \citet{tong2020trajectorynet} simulates samples using CNFs with a specific dynamic optimal transport-induced regularization, and this leads to approximately solving a Schrödinger bridge problem where the volatility term is 0 and the reference is a Brownian motion. Therefore, similarly to what we discussed in the main text for \citet{chen2024deep}, all these methods suffer from the fact that it is as if they need a pre-defined reference measure. And it is very rare that scientists have access to all the information about the correct reference measure for their problems. Our proposed method directly addresses this challenge by moving beyond a fixed reference process to improve trajectory inference. We emphasize that our main contribution lies in the reference refinement step, which, in theory, could be integrated with the methods discussed here as well as Schrödinger bridge methods outlined in the main text. This is an exciting direction for future work.

\section{Implementation details}
\label{sec:implementation}

In this section we discuss the implementation details of our method.

\subsection{First projection at one consecutive pair}
\label{sec:first_projection_implementation}
In this step, we are interested in estimating $\totsteps-1$ pairs of forward and backward drifts, $\{\estimatedriftstep{i},\estimatebackwarddriftstep{i}\}_{\timeidx = 1}^{\totsteps-1}$, that are ideally one the time inversion of the other. 

In our implementation, we decided to use the work in \citet{Vargas2021}. In particular, we use Algorithm 2 in the paper\footnote{The code implementation is available at \href{https://github.com/franciscovargas/GP_Sinkhorn}{\text{https://github.com/franciscovargas/GPSinkhorn}}}. We use the available code as it is and stick to the authors' choice for most parameters. The only difference comes from the fact that in \citet{Vargas2021}, the authors have only access to two time steps, $t_0$ and $t_1$, such that $t_1-t_0 = 1$. The authors in the paper pick discretization step $\Delta t = 0.01$, and therefore the total number of steps in their algorithm is $N = 1 / \Delta t = 100$. In our method, we also pick $\Delta t = 0.01$, but we have multiple time intervals, not necessarily all of length 1. Therefore we have a different amount of discretized steps, $N_i=[(\timet_{\timeidx+1}-\timet_{\timeidx})/\Delta\timet]$, for each time interval $\timeidx$. 

Note that in \citet{Vargas2021} the authors suggested 5 iterations are enough for this algorithm to converge \citep[see also Fig.~4 of][]{Vargas2021}, but we decided to run their algorithm for 10 iterations, which is the default in the available code \citep{gpshinkhornrepo}.

\subsection{Additional Discussion on Alternative Methods for the First Projection}
\label{app:sb-methods-discussion}

In this section, we include a discussion about additional methods that can be used for the first projection in our algorithm.

\paragraph{Iterative Markovian Fitting and Other Schr\"odinger Bridge Approaches.} A number of recent papers propose methods for fitting Markovian processes via iterative algorithms or by formulating SB problems directly in a forward-backward manner \citep{peluchetti2023diffusion, shi2023diffusion, gushchin2024light, Vargas2021, wang2021deep, Caluya2022, gu2024partially}. As discussed in the main text, our method is not restricted to the particular SB solver of \citet{Vargas2021} but can in principle incorporate \emph{any} algorithm that: (1) accepts a non-Brownian reference measure, and (2) outputs forward and backward drifts for each time interval. However, many of the existing SB methods mentioned above focus on Brownian-motion-based references (e.g., in image generation settings) and hard-code assumptions about isotropic diffusion or gradient drifts, making them less straightforward to incorporate directly without extensive modifications. Nonetheless, the underlying mathematical framework in most of these works can be adapted if one can suitably generalize beyond Brownian motion. In particular, the fast approach in \citet{gushchin2024light} emphasizes efficiency in solving Schr\"odinger Bridge problems with Brownian reference. Incorporating non-Brownian references would likely require re-deriving or extending its energy formulation to account for general SDE dynamics or a more flexible potential term. And \citet{de2021diffusion, peluchetti2023diffusion} also focus on bridging with Brownian motion in the context of generative modeling. Their code often encodes Brownian assumptions, so switching to a generic reference might demand non-trivial changes. Theoretically, however, these methods could still fit within our framework as long as one can compute or approximate drifts with respect to the new reference measure. We emphasize that our proposed reference refinement step (Step~2 in the main text) is computationally light, typically involving only a simple least-squares fit. Most of the runtime in our pipeline arises from solving the SB problem for consecutive pairs (Step~1). Therefore, incorporating a faster SB solver (e.g., via \citet{gushchin2024light}) can speed up the entire algorithm, though our iterative multi-marginal approach will necessarily be somewhat slower than a \emph{single} multi-marginal SB approach using the exact same solver.

\paragraph{Action-Matching.} Another interesting alternative for solving the first projection is the action-matching algorithm \citep{neklyudov2023action}.
This approach typically models particle trajectories via an ODE, with randomness only in the initial conditions, rather than a noisy SDE. By contrast, our setting presumes a genuinely stochastic process (i.e., the noise is part of the system). Nevertheless, both frameworks share a variational flavor: the action-matching objective for an ODE can be seen as minimizing energy consumption along a path, akin to the stochastic control formulation of SB problems. If one wished to extend action-matching to incorporate a non-Brownian reference process, the approach would involve adjusting the relevant energy functional to reflect that reference. In principle, our multi-marginal extension, where the reference is iteratively updated, could integrate with such an approach, subject to the difference that the baseline system is an ODE rather than an SDE.

\paragraph{Entropic Optimal Transport and Other Optimal Transport Solvers.} The Schr\"odinger Bridge (SB) problem with a Brownian reference is closely related to \emph{entropic-regularized} optimal transport. Consequently, algorithms such as SCONES \citep{daniels2021score} and EGNOT \citep{mokrov2023energy} could, in principle, serve as substitutes for the first projection step in our method, provided one can formulate each pairwise SB step as an entropic OT problem. However, these solvers typically assume a Brownian reference by design and become more complex to apply when the reference measure is generalized or only available through a black-box function. In such cases, it would be necessary to re-derive the underlying operator or cost function to account for nontrivial drift and diffusion. We view this as an exciting avenue for future research: if entropic OT solvers can be extended to arbitrary reference measures, then they could seamlessly replace the SB subroutine in our pipeline.

\subsection{Sampling routine}
\label{sec:sampling-routine}
Assume that we have access to $\estimatedrift$ and $\estimatebkwdrift$. In practice, as discussed at the end of the ``Multi-marginal Schrödinger bridges" paragraph in \cref{sec:setup}, these two drifts are constructed by concatenating $\totsteps-1$ pairs of forward and backward drifts, denoted as $\{\estimatedriftstep{i},\estimatebackwarddriftstep{i}\}_{\timeidx = 1}^{\totsteps-1}$. Here we are interested in generating trajectories from the underlying distribution. More precisely, we are interested in approximating one trajectory $\simulatedresponseat{\timet}^{ (\timeidx,\sampleidx_{\timeidx}) }$, ${0< \timet\le \timestep{\totsteps}}$, for each observed sample $\obsat{\timestep{\timeidx}}^{\sampleidx_{\timeidx}}$. 

We can do this by numerically solving the SDEs associated to these drifts using a standard Euler-Maruyama approach \citep[][Chapter 3.4]{sarkka2019applied} with sampling rate $\Delta\timet$. We can obtain $\simulatedresponseat{\timet}^{ (\timeidx,\sampleidx_{\timeidx}) }$ for ${\timestep{\timeidx}< \timet\le \timestep{\totsteps}}$, by starting from $\obsat{\timestep{\timeidx}}^{\sampleidx_{\timeidx}}$ and simulating forward, with discrete step $\Delta\timet$, the SDEs with drifts $\{\estimatedriftstep{\timeidx}, \estimatedriftstep{\timeidx+1},\ldots,\estimatedriftstep{\totsteps-1}\}$. We start at $\timestep{\timeidx}$ and use $\estimatedriftstep{\timeidx}$ to simulate forward  $\obsat{\timestep{\timeidx}}^{\sampleidx_{\timeidx}}$ to obtain the trajectory at discrete times $\timet_{\timeidx}+\Delta t, \timet_{\timeidx}+2\Delta t, \ldots$ . We do this for $L_\timeidx$ steps, where $L_\timeidx$ is the smallest integer such that $\timet_{\timeidx}+L_\timeidx \Delta t \ge t_{i + 1}$. That is, we use $\estimatedriftstep{\timeidx}$ as long as we are diffusing over the time interval $[\timestep{\timeidx}, \timestep{\timeidx+1}]$. As soon as our discrete solving routine gives us an observation beyond time $\timestep{\timeidx+1}$, we stop using $\estimatedriftstep{\timeidx}$ and start using $\estimatedriftstep{\timeidx+1}$. In particular, we start using $\estimatedriftstep{\timeidx+1}$ from $\simulatedresponseat{\timet_{\timeidx}+L_\timeidx \Delta t}^{(\timeidx,\sampleidx_{\timeidx})}$, and simulate this forward for $L_{\timeidx+1}$ steps, until $\timet_{\timeidx}+(L_\timeidx + L_{i+1})\Delta t \ge \timet_{\timeidx + 2}$. And we proceed in this way until we have simulated all the way forward until $\timestep{\totsteps}$. With this routine, it might happen that in the very last step we obtain $\simulatedresponseat{\timet_{\timeidx}+\sum_{j=i}^{\totsteps-1} L_j \Delta t}^{(\timeidx,\sampleidx_{\timeidx})}$ with $\timet_{\timeidx}+\sum_{j=i}^{\totsteps-1} L_j \Delta t \ge \timet_{\totsteps}$. Since we are assuming A1 and A2, and the trajectories are continuous, we do not expect the trajectory to vary a lot in the time past $\timet_{\totsteps}$, so we just consider the trajectory up to  $\simulatedresponseat{\timet_{\timeidx}+\sum_{j=i}^{\totsteps-1} L_j \Delta t}^{(\timeidx,\sampleidx_{\timeidx})}$. In a very similar manner, we obtain $\simulatedresponseat{\timet}^{ (\timeidx,\sampleidx_{\timeidx}) }$ for $0\le \timet\le \timestep{\timeidx}$, by starting from $\obsat{\timestep{\timeidx}}^{\sampleidx_{\timeidx}}$ and simulating backward, with discrete step $\Delta\timet$, the SDEs with drifts $\{\estimatebackwarddriftstep{\timeidx}, \estimatebackwarddriftstep{\timeidx-1},\ldots,\estimatebackwarddriftstep{1}\}$. We can then obtain a full trajectory for $\obsat{\timestep{\timeidx}}^{\sampleidx_{\timeidx}}$ by concatenating the backward and forward simulated trajectories. We denote this by $\simulatedresponseat{0 \le \timet\le \timestep{\totsteps}}^{(i,\sampleidx_{\timeidx}) }$. If we do this for all the observations, we obtain the set of trajectories $\{\simulatedresponseat{0\le \timet\le \timestep{\totsteps}}^{(\timeidx,\sampleidx_{\timeidx})}\}_{\timeidx,\sampleidx_{\timeidx}}$.

\subsection{Second projection}
\label{app:inverseproj}
Here we describe the subroutine we denoted as MLEfit in Algorithm~\ref{algo:sb-irr-algo}. 

We first observe that the inverse projection, \cref{eq:second-proj}, is equivalent to maximizing the cross entropy between $\refden_{\refdrift}$ and $\bridge_{\drift^{(\iteralgo)}}$

\begin{align*}
    \argmin_{\refdrift: \refden_{\refdrift}\in\modelfamily} \, \KL(\bridge_{\drift^{(\iteralgo)}} \| \refden_{\refdrift}) &= \argmin_{\refdrift: \refden_{\refdrift}\in\modelfamily} \E_{\bridge_{\drift^{(\iteralgo)}}}\log \frac{\bridge_{\drift^{(\iteralgo)}}}{\refden_{\refdrift}} \\
    &= \argmin_{\refdrift: \refden_{\refdrift}\in\modelfamily}\left[\E_{\bridge_{\drift^{(\iteralgo)}}}\log \bridge_{\drift^{(\iteralgo)}}-\E_{\bridge_{\drift^{(\iteralgo)}}}\log \refden_{\refdrift}\right] = \argmax_{\refdrift: \refden_{\refdrift}\in\modelfamily} \E_{\bridge_{\drift^{(\iteralgo)}}}\log \refden_{\refdrift}
\end{align*}

That is, the second projection can be viewed as the expected log-likelihood when true data is from $\bridge_{\drift^{(\iteralgo)}}$ and our model is $\refden_{\refdrift}$. We do not have access to the full distribution $\bridge_{\drift^{(\iteralgo)}}$, but only to the time discretized samples obtained from the previous sampling routine, $\{\simulatedresponseat{0\le \timet\le \timestep{\totsteps}}^{(\timeidx,\sampleidx_{\timeidx})}\}_{\timeidx,\sampleidx_{\timeidx}}$. Therefore, we can approximate the expectation with a sample mean.

To do so, observe that the likelihood of these time discretized samples under model $\refden_{\refdrift}$ can be calculated by an autoregressive process. Indeed, consider the estimated trajectory at some time $\timet_{\timeidx}$, $\simulatedresponseat{\timet_{\timeidx}}^{ (\timeidx,\sampleidx_{\timeidx}) }$. Under the SDE model $\refden_{\refdrift}$ with drift $\refdrift$ and sampling rate $\Delta\timet$, the next evaluation for the estimated trajectory, $\simulatedresponseat{\timet_{\timeidx} + \Delta\timet}^{ (\timeidx,\sampleidx_{\timeidx}) }$ is obtained by sampling from a Gaussian $\simulatedresponseat{\timet_{\timeidx} + \Delta\timet}^{ (\timeidx,\sampleidx_{\timeidx}) }|\simulatedresponseat{\timet_{\timeidx}}^{ (\timeidx,\sampleidx_{\timeidx}) }\sim \gN(\simulatedresponseat{\timet_{\timeidx}}^{ (\timeidx,\sampleidx_{\timeidx}) }+\refdrift(\simulatedresponseat{\timet_{\timeidx}}^{ (\timeidx,\sampleidx_{\timeidx}) }, \Delta\timet),\volatility \Delta\timet)$. Therefore, each point in the trajectory is dependent only on the immediate previous point, with the next value being a normally distributed variable centered around the previous value adjusted by the drift term 
$\refdrift$, with variance $\volatility \Delta t$. This process can thus be seen as a Markov chain with Gaussian transition kernel. 

Therefore, we can approximate $\E_{\bridge_{\drift^{(\iteralgo)}}}\log \refden_{\refdrift}$ by the sample mean
\begin{equation}
\label{eq:second_proj_L2}
\begin{aligned}
    \E_{\bridge_{\drift^{(\iteralgo)}}}\log \refden_{\refdrift} \approx& \frac{1}{\sum_{\timeidx=1}^{\totsteps}N_{\timeidx}}\frac{1}{\sum_{\timeidx} L_{\timeidx}}\sum_{\timeidx=1}^{\totsteps}\sum_{n_{\timeidx}=1}^{N_{\timeidx}} \sum_{\discstep=1}^{\totdiscstep_{\timeidx}-1} -\frac{1}{2\volatility\Delta\timet} ||(\simulatedresponseat{(\ell+1) \Delta \timet}^{ (\timeidx,\sampleidx_{\timeidx}) }-\simulatedresponseat{\ell \Delta \timet}^{ (\timeidx,\sampleidx_{\timeidx}) })-\refdrift(\simulatedresponseat{\ell \Delta \timet}^{ (\timeidx,\sampleidx_{\timeidx}) }, \discstep\Delta\timet)||_2^2 -\frac{1}{2}\log(\volatility \Delta\timet)
\end{aligned}
\end{equation}

And so to find the new optimal reference drift $\estimaterefdriftstep{}$ it is enough to solve the following minimization problem:

\begin{align*}
    \estimaterefdriftstep{}=\argmin_{\refdrift: \refden_{\refdrift}\in\modelfamily} \frac{1}{\sum_{\timeidx=1}^{\totsteps}N_{\timeidx}}\frac{1}{\sum_{\timeidx} L_{\timeidx}}\sum_{\timeidx=1}^{\totsteps}\sum_{n_{\timeidx}=1}^{N_{\timeidx}} \sum_{\discstep=0}^{\totdiscstep_{\timeidx}-1} \frac{1}{2\volatility\Delta\timet} ||(\simulatedresponseat{(\ell+1) \Delta \timet}^{ (\timeidx,\sampleidx_{\timeidx}) }-\simulatedresponseat{\ell \Delta \timet}^{ (\timeidx,\sampleidx_{\timeidx}) })-\refdrift(\simulatedresponseat{\ell \Delta \timet}^{ (\timeidx,\sampleidx_{\timeidx}) }, \discstep\Delta\timet)||_2^2
\end{align*}

\subsection{Code, hyperparameters, reference choice and training for our method}
\label{app:train-hyperparams-ref}

The code, data, and instructions needed to reproduce the main experimental results are provided at 
\ifthenelse{\boolean{accepted}}{\href{https://github.com/YunyiShen/SB-Iterative-Reference-Refinement}{this link}.}{\href{https://github.com/YunyiShen/SB-Iterative-Reference-Refinement}{this link}.}

We can solve the minimization problem described in \cref{app:inverseproj} by any standard nonlinear least squares routine. In our implementation, for each experiment we parametrize the reference family of drifts as a neural network (more precisely, a \texttt{nn.Module} in PyTorch), and train this neural network to predict all the finite differences $\simulatedresponseat{(\ell+1) \Delta \timet}^{ (\timeidx,\sampleidx_{\timeidx}) }-\simulatedresponseat{\ell \Delta \timet}^{ (\timeidx,\sampleidx_{\timeidx}) }$ using an MSE loss and gradient descent.  We provide more details for the reference family choice (and corresponding neural architecture) for each experiment in \cref{app:LV_implementation} (Lotka-Volterra), \cref{app:repr_implementation} (repressilator), \cref{app:eb_implementation} (single cell) and  \cref{app:GoM-app-exp-details} (ocean current). During training, the learning rate is set to 0.01 and we consider 50 epochs. These hyperparameter values were determined by a grid search on combinations of learning rates $\{0.05, 0.01, 0.005, 0.001\}$ and epochs of $\{20, 50\}$. The best choice was determined by visually inspecting generated trajectories passing through training data and quantitatively inspecting the loss in the second projection~\cref{eq:second_proj_L2}. We used the parameters that 1) the learned trajectories passes training data and 2) achieve the smallest second projection loss. 

We iterate our algorithm 10 times ($K = 10$) to refine the model, as we find empirically that this number best balances efficiency and accuracy across all experiments --- further iterations yield minimal additional gains. Appendix~\ref{app:multisteps-app} provides supporting results, illustrating how the test metric evolves throughout these refinement steps.

\subsection{\vanilla{}}
\label{sec:vanilla-app}
The \vanilla{} algorithm can be viewed as using the first step in our algorithm (where we solve the first KL projection) one time. Therefore, also for this method we decided to stick to Algorithm 2 by \citet{Vargas2021}\footnote{The code implementation available at \href{https://github.com/franciscovargas/GP_Sinkhorn}{https://github.com/franciscovargas/GPSinkhorn}}. We apply their code exactly as explained in \cref{sec:first_projection_implementation}.  

\subsection{Deep Momentum-SB}
\label{sec:dmsb-app}
For this additional baseline, we use the code provided by the authors of the paper at \href{https://github.com/TianrongChen/DMSB.}{https://github.com/TianrongChen/DMSB}. The code is released under MIT license \citep{dmsb}. We use their code with default parameters, as in the \texttt{"gmm"} experiment. Since the method seems to produce non-realistic trajectories with these settings, we performed a small grid search over the (\texttt{v\_scale}) parameter. We try values $\{0.1,0.01,0.001,0.0001,0.00001\}$ and find that we obtain the best trajectories --- both visually and in terms of EMD --- with \texttt{v\_scale}=0.01. This is aligned with the authors' choice in the \texttt{"RNA"} experiment. We also try to run our experiments using all the default parameters in the \texttt{"RNA"} experiment, but find out that the hybrid combination of \texttt{"gmm"} hyperparameters and $\texttt{v\_scale}=0.01$ is what works best both visually and in terms of EMD.

\subsection{TrajectoryNet}
\label{app:trajnetdetails}
For this baseline, we use the algorithm implemented in the Python package \texttt{TrajectoryNet} \citep{trajnet}. This package is licensed under Yale Non-Commercial License. We use the default parameter settings as specified in \citet{tong2020trajectorynet}. The algorithm returns trajectories starting from the last time step, going backwards to the first time step. We use these trajectories in our comparison.

\section{Iterative projections proof details}
\label{app:iter_proj_theory}

In this section we provide details on how to prove \cref{prop:iter-kl} in the main text. This result follows using standard theory of iterative projections \citep{csiszar1975divergence, csiszar2004information, benamou2015iterative}. 
In particular, the convergence results in our setting are derived by adapting Theorem 5.3 from  \citet{csiszar2004information}, originally formulated for families of discrete distributions. We have numbered this as \cref{thm:csiszar5.3}, and modified it to apply to density functions rather than probability mass functions. Once we have stated \cref{thm:csiszar5.3}, to prove \cref{prop:iter-kl} we need to apply this adaptation to the specific case where the divergence of interest is the KL divergence between SDEs. To do so, we need to verify that the theorem's conditions on probability densities are satisfied by our SDEs. To facilitate this, we have established additional lemmas --- \cref{lemma:pyth}, \cref{lemma:rIproj}, and~\cref{lemma:bounded_density_ratio}. Using these lemmas, we can then apply \cref{thm:csiszar5.3} in the specific case of KL divergence between densities defined by SDEs, and establish \cref{prop:iter-kl}. Our proof relies on expressing the probability distribution of trajectories as a density with respect to the Wiener measure. We leave the analysis of finite-sample behavior to future work.

This appendix is organized as follows:
\begin{itemize}
    \item We first briefly discuss the existence and convexity of densities associated to SDEs in our setting --- this will be needed to state and prove the main results in the rest of \cref{app:iter_proj_theory}. We also prove that the family $\measurefamily$ in \cref{eq:measure-family} is always convex.
    \item We then state and prove \cref{thm:csiszar5.3}, our continuous case adaptation of Theorem 5.3 from \citet{csiszar2004information}.
    \item Next, we state and prove \cref{lemma:pyth} and \cref{lemma:rIproj}, continuous setting adaptations of Theorem 3.1 and 3.4 in \citet{csiszar2004information}, which will be needed to verify assumptions in \cref{thm:csiszar5.3} for the KL divergence.
    \item We prove \cref{lemma:bounded_density_ratio}, which is needed to verify the assumptions in \cref{lemma:pyth}, \cref{lemma:rIproj}, and \cref{thm:csiszar5.3} for the KL divergence between densities associated to SDEs.
    \item And finally, we put everything together and prove \cref{prop:iter-kl}.
\end{itemize}

We can now define convexity in our setting. 

\subsection{Existence of densities and convexity} 
\label{app:convex-densities}
In our work, we assume that all the SDEs share the same diffusion term $\sqrt{\volatility}dW_t$, with fixed $\volatility$. When we consider these SDEs over a finite time horizon, it is well-known that the probability distribution over trajectories admits a density with respect to the Wiener measure defined by a Brownian motion with volatility $\volatility$. This is because of Girsanov's theorem (see, e.g., Theorem 8.6.3 in \citet{oksendal2013stochastic}). This theorem guarantees that the probability distribution over trajectories induced by the solution to an SDE with a drift and a Brownian motion component can be transformed into a probability distribution over trajectories induced by just the Brownian motion (i.e., the Wiener measure) through a change of measure. Note that Girsanov's theorem holds true in this setting since assumptions A1 and A2 imply Novikov's condition, equation (8.6.7) in Theorem 8.6.3 in \citet{oksendal2013stochastic}.

Having established the existence of densities, we can now define convexity for a set densities. In particular, in the rest of our work we say that a set of densities, e.g. $\modelfamily$, is convex if for all densities $p,p'\in \modelfamily$ and all $\omega\in [0,1]$ we have $(1-\omega)\Tilde{p}+\omega \Tilde{p}'\in \modelfamily$. A simple example of a convex set of densities is a single-parameter family of mixtures of two known SDEs with the same volatility. For each particle, we flip a coin with probability $d \in [0,1]$. If the coin lands heads, the particle follows the first SDE; otherwise, it follows the second SDE. Since each SDE has a density, denoted $p_1$ and $p_2$, the mixture density is given by $d p_1 + (1 - d) p_2$. For any $\omega \in [0,1]$ and $d, d' \in [0,1]$, the convex combination of these mixtures is $\omega(d p_1 + (1 - d) p_2) + (1 - \omega)(d' p_1 + (1-d')p_2) = (\omega d + (1 - \omega) d') p_1 + \bigl[1 - (\omega d + (1 - \omega) d')\bigr] p_2$, which also lies in the family because $\omega d + (1 - \omega) d' \in [0,1]$. Hence, this collection of densities is convex.

Now we show that the family $\measurefamily$ in \cref{eq:measure-family} is always convex.
\begin{myLemma}[Convexity of $\measurefamily$]
\label{lemma:convexD}
$\measurefamily$ is convex, i.e.,  for any two $\bridge_{\modeldrift_1},\bridge_{\modeldrift_2}\in \measurefamily$, we have for all $\omega\in [0,1]$, the density $\bridge_{\modeldrift}:=\omega\bridge_{\modeldrift_1} + (1-\omega)\bridge_{\modeldrift_2}\in \measurefamily$.
\end{myLemma}
\begin{proof}
    Consider the marginal distribution induced by $\omega\bridge_{\modeldrift_1} + (1-\omega)\bridge_{\modeldrift_2}$ at $\timestep{\timeidx}, \bridge_{\modeldrift, \timestep{\timeidx}}$. This can be written as
    $$
    \bridge_{\modeldrift, \timestep{\timeidx}}=\omega\bridge_{\modeldrift_1,\timestep{\timeidx}} + (1-\omega)\bridge_{\modeldrift_2,\timestep{\timeidx}} = \omega\marginal{\timestep{\timeidx}}+(1-\omega)\marginal{\timestep{\timeidx}} = \marginal{\timestep{\timeidx}}
    $$
    where in the last equality we use the fact that $\bridge_{\modeldrift_1}$ and $\bridge_{\modeldrift_2}$ are in $\measurefamily$. Since this holds for all $\timeidx\in[I]$, we have that $\bridge_{\modeldrift}\in \measurefamily$ and therefore $\measurefamily$ is convex.
\end{proof}

\subsection{Continuous setting adaptation of Theorem 5.3, \texorpdfstring{\citet{csiszar2004information}}{}}

Now that we have clearly defined what we mean when we talk about convex families, we can state and prove \cref{thm:csiszar5.3}.

\begin{myTheorem}[Continuous setting adaptation of Theorem 5.3, \citet{csiszar2004information}]
\label{thm:csiszar5.3}

Let \( D(q, p) \) be a divergence between densities \( q \) and \( p \) from two convex families of (probability) densities \(\altmeasurefamily\) and \( \altmodelfamily \), respectively. Define \( p^*(q) := \arg\min_{p\in\altmodelfamily} D(q,p) \) and \( q^*(p) := \arg\min_{q\in \altmeasurefamily} D(q,p) \). Consider the iterative scheme given by \( q^{(k)} = q^*(p^{(k-1)}) \) and \( p^{(k)} = p^*(q^{(k)}) \). Assume the existence of a non-negative function \( \delta(q, p) \) such that the following conditions are satisfied:
\begin{itemize}
    \item[] \textbf{1. Three-Points Property:} For all \( q\in \altmeasurefamily \) and \( p\in\altmodelfamily \),
    \begin{equation}
        \delta(q, q^*(p)) + D(q^*(p), p) \leq D(q, p).
        \label{eq:three_point_property}
    \end{equation}
    
    \item[] \textbf{2. Four-Points Property:} For all \( q' \in \altmeasurefamily \) and \( p' \in \altmodelfamily \), if \( \min_{p\in\altmodelfamily} D(q, p) < \infty \), then
    \begin{equation}
        D(q', p') + \delta(q', q) \geq D(q', p^*(q)).
        \label{eq:four_point_property}
    \end{equation}
    \item[] \textbf{3. Boundedness Property:} For all \( p\in\altmodelfamily \), if \( \min_{q\in \altmeasurefamily} D(q, p) < \infty \), then
    \begin{equation}
        \delta(q^*(p), q^{(1)}) < \infty.
        \label{eq:boundedness}
    \end{equation}
\end{itemize}
Then, if \( \min_{q\in \altmeasurefamily, p\in\altmodelfamily} D(q, p) < \infty \), the iteration converges to the infimum of \( D(q, p) \) over \( \altmeasurefamily \) and \( \altmodelfamily \):
\[
\lim_{k \to \infty} D(q^{(k)}, p^{(k)}) = \inf_{q\in \altmeasurefamily, p\in\altmodelfamily} D(q, p).
\]
\end{myTheorem}

\begin{proof}
First, from the Three-Points property we have that for all $q \in \altmeasurefamily$:
\[
\delta(q, q^{(k+1)}) + D(q^{(k+1)}, p^{(k)}) \leq D(q, p^{(k)}),
\]
where $q^{(k+1)} = q^*(p^{(k)})$. Then, using the Four-Points property, we get:
\[
D(q, p^{(k)}) \leq D(q, p) + \delta(q, q^{(k)}).
\]
for all $q\in\altmeasurefamily$ and $p\in\altmodelfamily$ because $p^{(k)} = p^{*}(q^{(k)})$. Combining these results yields:
\[
\delta(q, q^{(k+1)}) \leq D(q, p) - D(q^{(k+1)}, p^{(k)}) + \delta(q, q^{(k)}).
\]
This inequality shows the change in the discrepancy measure $\delta$ across iterations. By the definition of the iterative process, the sequence of divergences decreases monotonically:
\[
D(q^{(1)}, p^{(0)}) \geq D(q^{(1)}, p^{(1)}) \geq D(q^{(2)}, p^{(1)}) \geq \ldots
\]

Assume for contradiction that $\lim_{k \to \infty} D(q^{(k)}, p^{(k)}) \neq \inf_{q, p} D(q, p)$. Suppose there exists $p$ and $\epsilon > 0$ such that:
\[
D(q^{(k+1)}, p^{(n)}) > D(q^*(p), p) + \epsilon, \quad \text{for all } k.
\]
Applying the earlier derived inequality to $q = q^*(p)$, we find that for all $k$:
\[
\delta(q^*(p), q^{(k+1)}) \leq \delta(q^*(p), q^{(k)}) - \epsilon.
\]
Since $\delta(q^*(p), q^{(1)}) < \infty$, by assumption the Boundedness property, this recursion results in:
\[
\delta(q^*(p), q^{(k+1)}) < 0 \text{ for some } k,
\]
contradicting the non-negativity of $\delta$. Therefore, our initial assumption must be false, and we conclude:
\[
\lim_{k \to \infty} D(q^{(k)}, p^{(k)}) = \inf_{q, p} D(q, p).
\]
\end{proof}

\subsection{Continuous setting adaptation of Theorem 3.1, \texorpdfstring{\citet{csiszar2004information}}{}}

In order to apply this theorem to prove \cref{prop:iter-kl}, we need to show that the three properties (\cref{eq:three_point_property,eq:four_point_property,eq:boundedness}) are true when $D(q,p)=\KL(q,p)$, $\delta(q,p)=\KL(q,p)\ge 0$, and $p,q$ are densities.

To this end, we first state and prove \cref{lemma:pyth}, which is a continuous setting adaptation of Theorem 3.1 of \citet{csiszar2004information}. This establishes that  --- under one additional assumption on densities --- the three point property, \cref{eq:three_point_property}, is satisfied for $\delta(q,p)=D(q,p)=\KL(q||p)$.

\begin{myLemma}[Continuous setting adaptation of Theorem 3.1 of \citet{csiszar2004information}]
\label{lemma:pyth}
Let $\altmeasurefamily$ and $\altmodelfamily$ be two convex sets of densities with respect to a base measure $\basemeasure$. For any $q,q'\in \altmeasurefamily$ and $p\in \altmodelfamily$, assume $\E_{q}|\log(q'/p)| < \infty$. Define $q^*(p) := \argmin_{q\in \altmeasurefamily}\KL(q||p)$. Then, for any $q \in \altmeasurefamily$ and $p \in \altmodelfamily$:
$$\KL(q||p) \ge \KL(q||q^*) + \KL(q^*||p)$$
\end{myLemma}
\begin{proof}
Let $q_{\omega} = (1-\omega)q^* + \omega q \in \altmeasurefamily$ for $\omega \in [0,1]$. By the mean value theorem, $\exists \tilde{\omega} \in (0,1)$ such that:
$$0 \le \frac{1}{\omega} [\KL(q_{\omega}||p) - \KL(q^*||p)] = \frac{d}{d\omega} \KL(q_{\omega}||p) |_{\omega = \tilde{\omega}}$$
The lower bound is 0 as $q^*$ minimizes KL divergence. We now examine the derivative:
$$\frac{d}{d\omega} \KL(q_{\omega}||p) = \frac{d}{d\omega} \int q_{\omega} \log \frac{q_{\omega}}{p}\basediff$$
$q_{\omega} \log \frac{q_{\omega}}{p}$ is integrable by our initial assumption. To move the derivative inside the integral, we need to show the partial derivative is integrable. The partial derivative is:
$$\frac{\partial}{\partial \omega} q_{\omega} \log \frac{q_{\omega}}{p} = (q-q^*)(\log \frac{q_{\omega}}{p}+1)$$
We can bound this expression:
$$|(q-q^*)(\log \frac{q_{\omega}}{p}+1)| \le (q+q^*)|\log \frac{q_{\omega}}{p}+1|$$
Now we show this bound is integrable:
$$\int (q+q^*)|\log \frac{q_{\omega}}{p}+1|\basediff \le \E_{q} |\log \frac{q_{\omega}}{p}| + \E_{q^*} |\log \frac{q_{\omega}}{p}| + 2 < \infty$$
This integrability allows us to apply the Dominated Convergence Theorem (DCT):
$$\frac{d}{d\omega} \KL(q_{\omega}||p) = \int (q-q^*)\log \frac{q_{\omega}}{p}\basediff$$
We can now examine the limit as $\omega$ approaches 0. Again applying DCT:
$$\lim_{\omega \to 0}\frac{d}{d\omega} \KL(q_{\omega}||p) = \int (q-q^*)\log \frac{q^*}{p}\basediff \ge 0$$
The inequality holds because $q^*$ minimizes KL divergence. Finally, we expand this expression:
$$\int (q-q^*)\log \frac{q^*}{p}\basediff = \KL(q||p) - \KL(q||q^*) - \KL(q^*||p) \ge 0$$
This final step proves the stated inequality, completing our proof.
\end{proof}

\subsection{Continuous setting adaptation of Theorem 3.4, \texorpdfstring{\citet{csiszar2004information}}{}}

Next we show \cref{lemma:rIproj}, which is a continuous setting version of Theorem 3.4 of \citet{csiszar2004information}. This establishes that --- under the same additional assumption on densities that we have for \cref{lemma:pyth} --- the four-point property, \cref{eq:four_point_property}, is satisfied for $\delta(q,p)=D(q,p)=\KL(q||p)$.

\begin{myLemma}[modified version of theorem 3.4 of \citet{csiszar2004information}]
\label{lemma:rIproj}
    Let $\altmeasurefamily$ and $\altmodelfamily$ be two convex sets of densities with respect to some base measure $\basemeasure$. Assume for all $q \in \altmeasurefamily$ and $p, p' \in \altmodelfamily$, we have $\E_{q} [p/p'] < \infty$. Then, $p^* \in \altmodelfamily$ is the minimizer of $\KL(q||p)$ over $p \in \altmodelfamily$ if and only if for all $q' \in \altmeasurefamily$ and $p' \in \altmodelfamily$:
$$\KL(q'||p') + \KL(q'||q) \ge \KL(q'||p^*)$$
\end{myLemma}
\begin{proof}

\textbf{"If" part}: Take $q'=q$. The inequality directly implies $p^*$ is the minimizer.

\textbf{"Only if" part:} We claim that 
$$\int q(1-\frac{p'}{p^*})\basediff \ge 0$$. 
If this holds, then $1-\int \frac{qp'}{p^*}\basediff \ge 0$, and 
$$\int q'(1-\frac{qp'}{q'p^*})\basediff \ge 0$$
Using the inequality $\log(1/x) \ge 1-x$, we have:
$$\KL(q'||p') + \KL(q'||q) - \KL(q'||p^*) = \int q'\log\frac{q'p^*}{qp'}\basediff \ge \int q'(1-\frac{qp'}{q'p^*})\basediff \ge 0$$
This proves the main result. Thus, it suffices to prove $\int q(1-\frac{p'}{p^*})\basediff \ge 0$.

To prove this claim, set $p_{\omega} = (1-\omega)p^* + \omega p' \in \altmodelfamily$ for $\omega \in [0,1]$. Since $p^*$ is a minimizer, by the mean value theorem, $\exists \tilde\omega \in (0,1)$ such that:
$$0 \le \frac{1}{\omega}(\KL(q||p_\omega) - \KL(q||p^*)) = \frac{d}{d\omega} \KL(q||p_\omega) |_{\omega = \tilde\omega}$$
To move the derivative inside the integral, we check if it's absolutely integrable:
$$\frac{d}{d\omega} q\log\frac{q}{(1-\omega)p^* + \omega p'} = q\frac{p^*-p'}{(1-\omega)p^* + \omega p'}$$
We show it has finite integral:
$$\int |q\frac{p^*-p'}{(1-\omega)p^* + \omega p'}|\basediff \le \E_{q} \frac{p^*}{(1-\omega)p^* + \omega p'} + \E_{q} \frac{p'}{(1-\omega)p^* + \omega p'} < \infty$$
This is finite by our assumption, as $(1-\omega)p^* + \omega p' \in \altmodelfamily$.

Taking the limit as $\omega \to 0$:
$$0 \le \lim_{\omega\to 0} \frac{d}{d\omega} \KL(q||p_\omega) = \lim_{\omega\to 0} \int q\frac{p^*-p'}{(1-\omega)p^* + \omega p'}\basediff$$
By the Dominated Convergence Theorem, we can move the limit inside the integral:
$$0 \le \int q\lim_{\omega\to 0} \frac{p^*-p'}{(1-\omega)p^* + \omega p'} \basediff= \int q(1-\frac{p'}{p^*})\basediff$$

This completes the proof of our claim and thus the lemma.
\end{proof}

\subsection{Expectation of density ratios}

Next, we state and prove \cref{lemma:bounded_density_ratio}. Informally, this lemma says that --- in our specific SDE setting --- the assumption on densities needed in \cref{lemma:pyth} and \cref{lemma:rIproj} is satisfied. The Boundedness Property, \cref{eq:boundedness}, in \cref{thm:csiszar5.3} is also a direct consequence of this lemma.

\begin{myLemma}[Expectation of density ratio]
\label{lemma:bounded_density_ratio}

Consider three SDEs that share the same diffusion term \(\sqrt{\volatility}\de\brownianm\) and have different drift terms \(\drift_{p}(\cdot,\cdot)\), \(\drift_{q}(\cdot,\cdot)\), and \(\drift_{q'}(\cdot,\cdot)\). Let \(p\), \(q\), and \(q'\) be the corresponding densities over the Wiener measure.

The SDEs are defined as follows:
\[
\begin{aligned}
    &\text{For density } p: &&\de\responseat{\timet} = \drift_{p}(\responseat{\timet},t)\de t + \sqrt{\volatility}\de\brownianm, \\
    &\text{For density } q: &&\de\responseat{\timet} = \drift_{q}(\responseat{\timet},t)\de t + \sqrt{\volatility}\de\brownianm, \\
    &\text{For density } q': &&\de\responseat{\timet} = \drift_{q'}(\responseat{\timet},t)\de t + \sqrt{\volatility}\de\brownianm.
\end{aligned}
\]

Assume that the time horizon is finite, i.e., \(t \in [0, T]\), and that all drift terms are bounded in the \(L_\infty\) norm. Specifically, there exists a constant \(C\) such that
\[
\|\drift_{i}(x,t)\|_{\infty} < C < \infty \quad \text{for } i \in \{p, q, q'\} \text{ and for all } x, t.
\]

Under these conditions, the density ratio between the processes is bounded. In particular, the following expectations are finite:
\begin{enumerate}
    \item The expected log density ratio: \(\mathbb{E}_{q} \left[ \left| \log \frac{q'}{p} \right| \right] < \infty\). (Assumption in \cref{lemma:pyth})
    \item The expected density ratio: \(\mathbb{E}_{p} \left[ \frac{q}{q'} \right] < \infty\). (Assumption in \cref{lemma:rIproj})
    \item The KL divergence between q and p: \(\mathbb{E}_{q} \left[ \frac{q}{p} \right] < \infty\) (Boundedness Property, \cref{eq:boundedness})
\end{enumerate}
\end{myLemma}

\begin{proof}
We first show that the expected density ratio is finite. To this end, note first that since the diffusion terms \(\sqrt{\volatility}\de\brownianm\) are shared among the SDEs, the three SDEs define distributions over trajectories that admit densities over a Wiener measure. By direct application of Girsanov's theorem \citep[see e.g.,][]{kailath1971structure}, the density ratio between the two measures on a sample path \(\responseat{\timet}\) is given by:
\[
\frac{q}{q'} = \exp\left[\frac{1}{\volatility} \int_0^T (\drift_{q}(\responseat{\timet}, t) - \drift_{q'}(\responseat{\timet}, t))^{\top} \de\brownianm - \frac{1}{2\volatility} \int_0^T \|\drift_{q}(\responseat{\timet}, t) - \drift_{q'}(\responseat{\timet}, t)\|_2^2 \de t \right].
\]

Next, we consider the expectation of this density ratio when trajectories are from another distribution over trajectories defined by the SDE with density \(p\). This expectation is:
\[
\begin{aligned}
    \E_{p}\left[\frac{q}{q'}\right] &= \E_{p}\left[\exp\left(\frac{1}{\volatility} \int_0^T (\drift_{q}(\responseat{\timet}, t) - \drift_{q'}(\responseat{\timet}, t))^\top \de\brownianm - \frac{1}{2\volatility} \int_0^T \|\drift_{q}(\responseat{\timet}, t) - \drift_{q'}(\responseat{\timet}, t)\|_2^2 \de t\right)\right].
\end{aligned}
\]

Using the assumption that the drift terms are bounded, i.e., \(\|\drift_{i}(x, t)\|_\infty < C < \infty\) for \(i = p, q, q'\), we get:
\[
\begin{aligned}
    \E_{p}\left[\frac{q}{q'}\right] &\le \E_{p}\left[\exp\left(\frac{1}{\volatility} \int_0^T 2C \bm{1}^\top \de\brownianm + \frac{1}{2\volatility} \int_0^T 4dC^2 \de t\right)\right] \\
    &= \exp\left(\frac{2dC^2 T}{\volatility}\right) \E_{p}\left[\exp\left(\frac{2C}{\volatility} \int_0^T \bm{1}^\top \de\brownianm\right)\right].
\end{aligned}
\]

The last factor, \(\exp\left(\frac{2C}{\volatility} \int_0^T \bm{1}^\top \de\brownianm\right)\), is a log-normal random variable, which has a finite expectation. Therefore:
\[
\E_{p}\left[\frac{q}{q'}\right] < \infty.
\]

For the first statement regarding the expected log density ratio, we proceed in a similar manner. We need to show:
\[
\begin{aligned}
    \E_{q}\left[\left|\log \frac{q'}{p}\right|\right] &= \E_{q}\left[\left| \int_0^T (\drift_{q'}(\responseat{\timet}, t) - \drift_{p}(\responseat{\timet}, t))^\top \de\brownianm - \frac{1}{2\volatility} \int_0^T \|\drift_{q'}(\responseat{\timet}, t) - \drift_{p}(\responseat{\timet}, t)\|_2^2 \de t \right|\right] < \infty
\end{aligned}
\]

Using the boundedness of the drift terms, we get:
\[
\begin{aligned}
    \E_{q}\left[\left|\log \frac{q'}{p}\right|\right] &\le \E_{q}\left[\left|\int_0^T \left|\drift_{q'}(\responseat{\timet}, t) - \drift_{p}(\responseat{\timet}, t)\right|^\top \de\brownianm\right| + \frac{1}{2\volatility} \int_0^T 4dC^2 \de t \right] \\
    &\le \E_{q}\left[\left|\int_0^T 2C \bm{1}^\top \de\brownianm\right|\right] + \frac{2dC^2 T}{\volatility}.
\end{aligned}
\]

The first term is finite since it is a Gaussian random variable, and the second term is finite by assumption that \(C < \infty\). Thus:
\[
\E_{q}\left[\left|\log \frac{q'}{p}\right|\right] < \infty.
\]

Finally, considering the case \(q = q'\), the KL divergence between \(q\) and \(p\) is finite under our assumption:
\[
\E_{q}\left[\log \frac{q}{p}\right] \le \E_{q}\left[\left|\log \frac{q}{p}\right|\right] < \infty.
\]

This completes the proof.
\end{proof}

\subsection{Proof of \texorpdfstring{\cref{prop:iter-kl}}{}}

We now have all the pieces needed to apply \cref{thm:csiszar5.3} in the specific case where $\delta(q,p)=D(q,p)=\KL(q||p)$, to restate and prove \cref{prop:iter-kl}.

\iterkl*

\begin{proof}
We show that this result follows directly from \cref{thm:csiszar5.3} by verifying all conditions. The convexity of $\measurefamily$ follows from \cref{lemma:convexD}. Consider $ q = \bridge_{\modeldrift}$, $p = \refden_{\refdrift}$. Then let the D divergence in \cref{thm:csiszar5.3} be the KL divergence, i.e., $D(q,p) = \KL(\bridge_{\modeldrift},\refden_{\refdrift})$ and $\delta(q,p) = \KL(\bridge_{\modeldrift},\refden_{\refdrift}) \ge 0$, where $p$ and $q$ are densities.

To apply \cref{thm:csiszar5.3} in this setting, we need to show that the three properties in \cref{thm:csiszar5.3} hold true:

\begin{enumerate}
    \item Three-point property (\cref{eq:three_point_property}):
    This property is established using \cref{lemma:pyth}. The conditions for \cref{lemma:pyth} are satisfied due to our assumption of bounded drift and \cref{lemma:bounded_density_ratio}.

    \item Four-point property (\cref{eq:four_point_property}):
    This property is established using \cref{lemma:rIproj}. The conditions for \cref{lemma:rIproj} are satisfied due to our assumption of bounded drift and \cref{lemma:bounded_density_ratio}.

    \item Boundedness assumption (\cref{eq:boundedness}):
    This assumption is directly established by \cref{lemma:bounded_density_ratio}.
\end{enumerate}

Given that all three required properties are satisfied, we can apply \cref{thm:csiszar5.3} to our setting, which completes the proof.
\end{proof}

\clearpage

\section{Further experimental details}
\label{app:further-experiments}
In this section we provide experimental details for each experiment, and report the results both visually and in terms of EMD and MMD.  We also showed how these metric change over 10 iterations of reference refinement. 

\subsection{Metric choice}
\label{app:metrics}
We choose to use two metrics that are commonly used in the literature to compare distributions, the Earth Mover Distance (EMD, \citet{rubner1998metric}, also known as Wasserstein-1) and the Maximum Mean Discrepancy (MMD, \citet{gretton2012kernel}). The EMD between two distributions $P$ and $Q$ can be defined by the maximization problem 
\begin{equation}
    \mathrm{EMD}(P,Q) = \sup_{||f||_L\le 1} \E_{x\sim P}f(x)-\E_{y\sim Q}f(y)
\end{equation} 
where $||f||_L\le 1$ constraints the function $f$ to be 1-Lipschitz continuous, i.e. $||\nabla f||\le 1$. The EMD intuitively measures the distance between two probability distributions by computing the minimum cost required to transform one distribution into the other. This metric allows us to quantify the similarity between the inferred and actual trajectories. This is a valid metric for our experiments, since if the inferred and actual trajectories are from the same distribution we expect it to converge to 0 (as the amount of trajectories goes to infinity). See \citet{fournier2015rate} for more details on this property. The metric can be calculated using the Sinkhorn algorithm. We use the implementation in the TrajectoryNet package. EMD is a widely used metric in the literature and remains useful in our setting because we work in low-dimensional spaces: our simulations are in 2D, and for the real data experiment, we reduce the data to the first five principal components. However, in higher-dimensional settings, EMD becomes less reliable due to the curse of dimensionality. Therefore, if a practitioner intends to apply our method to high-dimensional data, we recommend using MMD instead, as it is better suited for such cases.

The MMD between two distributions is defined in a very similar way. Instead of constraining the optimization problem to 1-Lipschitz continuous functions, we ask $f$ to be in a reproducing kernel Hilbert space $\mathcal{H}$ induced by a kernel function of choice.
\begin{equation}
    \mathrm{MMD}(P,Q) = \sup_{f\in\mathcal{H}} \E_{x\sim P}f(x)-\E_{y\sim Q}f(y)
\end{equation}

In our implementation, we pick the underlying kernel to be the Radial Basis Function (RBF) kernel with a length scale of 1 across all experiments. 

\subsection{Experiments tiebreaking details}
\label{app:tiebreaking}
In the tables across the paper, we highlight in green the method with lowest mean error over the restricted trajectories; we also highlight any other (restricted) method whose one-standard-deviation confidence interval overlaps the mean of the best (restricted) method. Separately, we use light blue to indicate any all-trajectory method whose one-standard-deviation confidence interval either overlaps with or falls below the mean of the best (restricted) method. 

\subsection{Lotka-Volterra}
\label{app:lv-app}
\subsubsection{Experiment setup}
For this experiment, we are interested in learning the dynamics of a stochastic Lotka-Volterra predator-prey model. The dynamics of the prey and predator populations are given by the following SDEs:
\begin{align}
\begin{split}
\label{eq:lv-family}
    dX &= \alpha X-\beta XY + 0.1dW_x\\
    dY &= \gamma XY-\delta Y + 0.1dW_y
\end{split}
\end{align}
where $[dW_x,dW_y]$ is a 2D Brownian motion.  

To obtain data, we fix the following parameters: $\alpha=1, \beta = 0.4, \gamma=0.1,\delta=0.4$. We start the dynamics at $X_0\sim U(5,5.1)$ and $Y_0\sim U(4,4.1)$, and simulate the SDEs for 10 instants of time. The length of each time interval is 1. We use Euler-Maruyama method to obtain the numeric solutions. We obtain 50 particles for each snapshot. We set $\Delta t = 0.01$ for every discretization step. We initialize the reference drift to 0, so that the initial reference SDE is a simple Brownian motion. 

\subsubsection{Reference family choice}
\label{app:LV_implementation}
 For this experiment, we have access to the data-generating process, as described in \cref{eq:lv-family}. Therefore, we select the reference family to be the set of SDEs that satisfy this system of equations, \cref{eq:lv-family}. We implement this reference family in practice as a PyTorch \texttt{nn.Module} with four scalar parameters, $\alpha, \beta, \gamma, \delta$. These parameters are to be learned during the optimization phase. Specifically, we use this module to evaluate $dX$ and $dY$, returning \texttt{torch.stack([dX, dY])} in the \texttt{nn.forward} method. The learning process involves optimizing the parameters using gradient descent, with a learning rate of 0.05 over 20 epochs, as determined by the grid search detailed in \cref{app:inverseproj}.

\subsubsection{Results} 
We provide a visual representation ot the trajectories in \cref{fig:LV-main} in the main text. In terms of EMD and MMD metrics, we show the results in \cref{tab:LV50_EMD_appendix} and \cref{tab:LV50_MMD}, respectively. We can see that our method is most of the times the best on the restricted set of trajectories (first four rows), and always at least as good as the other methods. When considering results over all possible trajectories (last two rows) our method is even better, especially for the last time step.

\begin{table*}[!ht]
    \centering
    \begin{tabular}{lcccc}
    \hline
        Method & EMD $t_2$ & EMD $t_4$ & EMD $t_6$ & EMD $t_8$ \\
    \hline
    \vanillaonetime & 0.59 $\pm$ 0.28 & 0.46 $\pm$ 0.073 & 0.40 $\pm$ 0.059 & \cellcolor{green!25}\textbf{0.70 $\pm$ 0.10} \\
         \dmsb & 0.49 $\pm$ 0.073 & 0.38 $\pm$ 0.12 & \cellcolor{green!25}\textbf{0.30 $\pm$ 0.11} & 0.84 $\pm$ 0.27\\
         \trajnet & 3.41 $\pm$ 0.050 & 1.27 $\pm$ 0.030 & 1.08 $\pm$ 0.040 & 3.99 $\pm$ 0.35 \\
         \oursonetime & \cellcolor{green!25}\textbf{0.22 $\pm$ 0.039} & \cellcolor{green!25}\textbf{0.16 $\pm$ 0.026} & \cellcolor{green!25}\textbf{0.31 $\pm$ 0.070} & \cellcolor{green!25}\textbf{0.74 $\pm$ 0.18} \\
    \hline
    \vanillaalltime & 0.59 $\pm$ 0.29 & 0.46 $\pm$ 0.074 & 0.38 $\pm$ 0.067 & \cellcolor{blue!10}\textbf{0.57 $\pm$ 0.072} \\
    \oursalltime & \cellcolor{blue!10}\textbf{0.21 $\pm$ 0.039} & \cellcolor{blue!10}\textbf{0.15 $\pm$ 0.021} & \cellcolor{blue!10}\textbf{0.30 $\pm$ 0.071} & \cellcolor{blue!10}\textbf{0.48 $\pm$ 0.055} \\
    \hline
    \end{tabular}
    \caption{Earth mover's distance (mean $\pm$ standard deviation) in four validation time points in Lotka-Volterra dataset with 50 particles each snapshot. Results were averaged over 10 seeds.}
    \label{tab:LV50_EMD_appendix}
\end{table*}

\begin{table*}[!ht]
    \centering
    \begin{tabular}{lcccc}
    \hline
        Method & MMD $t_2$ & MMD $t_4$ & MMD $t_6$ & MMD $t_8$ \\
    \hline
        \vanillaonetime & 2.73 \fpm 2.08 & 1.63 \fpm 0.54 & 0.68 \fpm 0.38 & \cellcolor{green!25}\textbf{0.78 \fpm 0.26} \\
         \dmsb & 2.03 $\pm$ 0.51 & 1.06 $\pm$ 0.69 & \cellcolor{green!25}\textbf{0.48 $\pm$ 0.36} & 1.26 $\pm$ 0.72\\
         \trajnet & 7.27 $\pm$ 0.11 & 6.33 $\pm$ 0.33 & 5.34 $\pm$ 0.20 & 6.35 $\pm$ 0.33 \\
         \oursonetime & \cellcolor{green!25}\textbf{0.40 \fpm 0.16} & \cellcolor{green!25}\textbf{0.068 \fpm 0.069} & \cellcolor{green!25}\textbf{0.34 \fpm 0.25} & \cellcolor{green!25}\textbf{0.80 \fpm 0.44} \\
    \hline
        \vanillaalltime & 2.75 $\pm$ 2.11 & 1.63 $\pm$ 0.55 & \cellcolor{blue!10}\textbf{0.67 $\pm$ 0.39} & \cellcolor{blue!10}\textbf{0.62 $\pm$ 0.23} \\    
        \oursalltime & \cellcolor{blue!10}\textbf{0.38 $\pm$ 0.16} & \cellcolor{blue!10}\textbf{0.05 $\pm$ 0.03} & \cellcolor{blue!10}\textbf{0.33 $\pm$ 0.25} & \cellcolor{blue!10}\textbf{0.29 $\pm$ 0.11} \\
    \hline
    \end{tabular}
    \caption{Maximum mean discrepancy (mean $\pm$ standard deviation) in four validation time points in Lotka-Volterra dataset with 50 particles. Results averaged over 10 seeds.}
    \label{tab:LV50_MMD}
\end{table*}

\clearpage

\subsection{Repressilator}
\label{app:repr-app}

\subsubsection{Experiment setup} 
The repressilator is a synthetic genetic regulatory network that functions as a biological oscillator, or a genetic clock. It was designed to exhibit regular, sustained oscillations in the concentration of its components. The repressilator system consists of a network of three genes that inhibit each other in a cyclic manner: each gene produces a protein that represses the next gene in the loop, with the last one repressing the first, forming a feedback loop.

We can model the dynamics of the repressilator using the following SDEs:
\begin{align}
\label{eq:repr-family}
    dX_1&=\frac{\beta}{1+(X_3/k)^n}-\gamma X_1+0.1dW_1 \nonumber \\
    dX_2&=\frac{\beta}{1+(X_1/k)^n}-\gamma X_2+0.1dW_2\\
    dX_3&=\frac{\beta}{1+(X_2/k)^n}-\gamma X_3+0.1dW_3 \nonumber
\end{align}
where $[dW_1,dW_2,dW_3]$ is a 3D Brownian motion, and the repressing behavior is quite clear from the drift equations. 

To obtain data, we fix the following parameters: $\beta = 10, n=3, k = 1, \gamma = 1$. We start the dynamics with initial distribution $X_1, X_2\sim U(1,1.1)$ and $X_3\sim U(2,2.1)$. We simulate the SDEs for 10 instants of time. At each time step, we take 50 samples. We use Euler-Maruyama method to obtain the numeric solutions. Also for this experiment, we set $\Delta t = 0.01$ for every discretization step. We initialize the reference drift to 0, so that the initial reference SDE is a simple Brownian motion.

\subsubsection{Reference family choice}
\label{app:repr_implementation}
For this experiment, we have access to the data-generating process, as described in \cref{eq:repr-family}. Therefore, we select the reference family to be the set of SDEs that satisfy this system of equations, \cref{eq:repr-family}. We implement this reference family in practice as a PyTorch \texttt{nn.Module} with four scalar parameters \(\beta, n, k, \gamma\) to be optimized by gradient descent. Specifically, we use this module to evaluate \(dX_1, dX_2\), and \(dX_3\), returning \texttt{torch.stack([$dX_1, dX_2, dX_3$])} in the \texttt{nn.forward} method. The learning process involves optimizing the parameters using gradient descent, with a learning rate of 0.05 over 20 epochs, as determined by the grid search detailed in \cref{app:inverseproj}.

\subsubsection{Results.} We provide a visual representation ot the trajectories in \cref{fig:repr-main} in the main text. In terms of EMD and MMD metrics, we show the results in \cref{tab:repres50_EMD_appendix} and \cref{tab:repres50_MMD}, respectively. We can see that our method is in general better than all the other methods on the restricted set of trajectories (first four rows), besides the last time point where \vanilla{} is the best. When considering results over all possible trajectories (last two rows), our method becomes even better at some time steps (especially $t_6$. \vanilla{} shows similar performance.

\begin{table*}[!ht]
    \centering
    \begin{tabular}{lccccc}
    \hline
        Method & EMD $t_2$ & EMD $t_4$ & EMD $t_6$ & EMD $t_8$ & EMD $t_{10}$ \\
    \hline
        \vanillaonetime& 1.87 $\pm$ 0.047 & 1.22 $\pm$ 0.12 & 1.33 $\pm$ 0.17 & \cellcolor{green!25}\textbf{1.20 $\pm$ 0.18} & \cellcolor{green!25}\textbf{1.16 $\pm$ 0.14} \\
         \dmsb & 1.46 $\pm$ 0.08 & 1.08 $\pm$ 0.37 & 3.00 $\pm$ 0.54 & 2.18 $\pm$ 0.41 & 2.54 $\pm$ 1.21\\
         \trajnet & 3.62 $\pm$ 0.05 & 2.86 $\pm$ 0.08 & 1.67 $\pm$ 0.07 & 3.45 $\pm$ 0.09 & 2.33 $\pm$ 0.08 \\
         \oursonetime & \cellcolor{green!25}\textbf{0.51 $\pm$ 0.11} & \cellcolor{green!25}\textbf{0.76 $\pm$ 0.10} & \cellcolor{green!25}\textbf{0.49 $\pm$ 0.10} & \cellcolor{green!25}\textbf{1.25 $\pm$ 0.27} & 2.18 $\pm$ 0.58 \\
    \hline
    \vanillaalltime & 1.87 $\pm$ 0.05 & 1.23 $\pm$ 0.10 & 1.27 $\pm$ 0.15 & \cellcolor{blue!10}\textbf{1.19 $\pm$ 0.13} & \cellcolor{blue!10}\textbf{1.16 $\pm$ 0.14} \\
    \oursalltime & \cellcolor{blue!10}\textbf{0.50 $\pm$ 0.11} & 0.87 $\pm$ 0.10 & \cellcolor{blue!10}\textbf{0.51 $\pm$ 0.09} & \cellcolor{blue!10}\textbf{0.61 $\pm$ 0.10} & 1.49 $\pm$ 0.20 \\
    \hline
    \end{tabular}
    \caption{Earth mover's distance (mean $\pm$ standard deviation) in five validation time points in repres50 dataset, our method performs better than the baseline SB.}
    \label{tab:repres50_EMD_appendix}
\end{table*}

\begin{table*}[!ht]
    \centering
    \begin{tabular}{lccccc}
    \hline
        Method & MMD $t_2$ & MMD $t_4$ & MMD $t_6$ & MMD $t_8$ & MMD $t_{10}$ \\
    \hline
        \vanillaonetime & 9.33 \fpm 0.059 & 5.52 \fpm 0.68 & 4.19 \fpm 0.68 & \cellcolor{green!25}\textbf{1.87 \fpm 0.62} & \cellcolor{green!25}\textbf{2.04 \fpm 0.50} \\
         \dmsb & 8.07 $\pm$ 0.29 & 4.37 $\pm$ 1.54 & 5.75 $\pm$ 0.37 & 3.65 $\pm$ 0.97 & 2.30 $\pm$ 0.83\\
         \trajnet & 6.86 $\pm$ 0.06 & 6.23 $\pm$ 0.13 & 4.56 $\pm$ 0.17 & 5.40 $\pm$ 0.33 & 4.61 $\pm$ 0.26 \\
         \oursonetime & \cellcolor{green!25}\textbf{1.97 \fpm 0.80} & \cellcolor{green!25}\textbf{2.70 \fpm 0.66} & \cellcolor{green!25}\textbf{0.60 \fpm 0.32} & \cellcolor{green!25}\textbf{1.40 \fpm 0.51} & \cellcolor{green!25}\textbf{2.20 \fpm 0.96}\\
    \hline
    \vanillaalltime & 9.33 $\pm$ 0.06 & 5.31 $\pm$ 0.72 & 3.87 $\pm$ 0.75 & \cellcolor{blue!10}\textbf{1.74 $\pm$ 0.69} & \cellcolor{blue!10}\textbf{1.70 $\pm$ 0.52} \\
    \oursalltime & \cellcolor{blue!10}\textbf{1.96 $\pm$ 0.80} & 3.34 $\pm$ 0.50 & \cellcolor{blue!10}\textbf{0.64 $\pm$ 0.24} & \cellcolor{blue!10}\textbf{0.51 $\pm$ 0.18} & \cellcolor{blue!10}\textbf{0.95 $\pm$ 0.23} \\
    \hline
    \end{tabular}
    \caption{Maximum mean discrepancy (mean $\pm$ standard deviation) in five validation time points in repres50 dataset.}
    \label{tab:repres50_MMD}
\end{table*}

\clearpage

\subsection{Gulf of Mexico vortex data}
\label{app:GoM-app}
\subsubsection{Experiment setup}
\label{app:GoM-app-exp-details}

\begin{figure}
    \centering
    \includegraphics[width=0.8\linewidth]{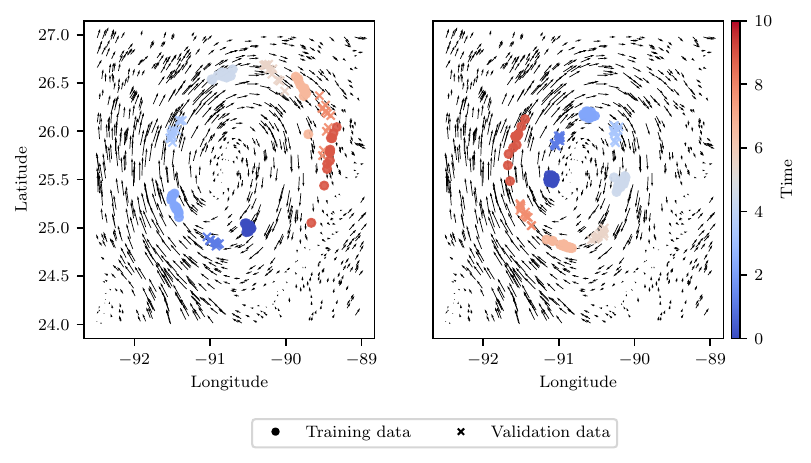}
    \caption{Our two datasets and the underlying ocean current dynamics from which we generated the data (black arrows). Left: \textit{Gulf of Mexico - big vortex} experiment. Right: \textit{Gulf of Mexico - small vortex} experiment. }
    \label{fig:gomdata}
\end{figure}

In this experiment, we test our method on real ocean-current data from the Gulf of Mexico. We use high-resolution (1 km) bathymetry data from a HYbrid Coordinate Ocean Model (HYCOM) reanalysis\footnote{Dataset available at \href{https://www.hycom.org/data/gomb0pt01/gom-reanalysis}{this link}.} \citep{panagiotis2014gulf}. This dataset was released by the US Department of Defense, and is thus of public domain. The dataset provides hourly ocean current velocity fields for the region extending from 98$^{\circ}$E to 77$^{\circ}$E in longitude and from 18$^{\circ}$N to 32$^{\circ}$N in latitude, covering every day since January 1\textsuperscript{st}, 2001. We focus on a specific time point, June 1st 2024 at 5pm at surface level, and a particular spatial region where a vortex is observed. By doing so, we obtain a unique velocity field with a known general behavior. Using this field, we generate particles that evolve according to the ocean currents while satisfying our modeling assumptions. Specifically, we select an initial location near the vortex and uniformly sample 1,000 initial positions within a small radius ($0.05$) around this point, representing the starting positions of 1,000 particles. We then evolve these particles over nine time steps using the ocean current velocity field. The time step size is $0.9$. Since the velocity field is defined on a fine grid, we approximate the velocity at each particle's position by using the velocity at the nearest grid point when the particle does not align exactly with a grid point. This approach simulates the particles' trajectories within the system. To create data that align with our modeling assumptions, we randomly assign each particle to one of the nine time steps, ensuring that (1) each particle is observed at only one time point, and (2) each time step has approximately the same number of particle observations. Specifically, we observe about 111 particles at each time step (since $1,000 \div 9 \approx 111$). This results in a dataset with sparse observations, where individual particle trajectories are not fully observed. We perform this data generation task twice, starting at two different locations around the vortex. See \cref{fig:gomdata} for a visual representation. We refer to the data in the left plot of \cref{fig:gomdata} as \textit{Gulf of Mexico - big vortex}. And to the data in the right plot as \textit{Gulf of Mexico - small vortex}. The goal of this experiment is to assess whether our method and the baseline methods can reconstruct the vortex from these two sparse particle datasets without access to complete individual trajectories.

\subsubsection{Reference family choice} 
\label{app:GoM-app-reference}

For this real data experiment, we do not have access to the data-generating process. However, we can leverage domain knowledge about the underlying ocean current vortex dynamics. Specifically, we model the dynamics of the vortex using the following SDEs:

\begin{align}
\label{eq:vortex-family}
    dX_1 & =\texttt{scale} \cdot \bigg((X_2 - X_2^{\text{center}}) \cdot \exp{(-\texttt{logyscale})}\bigg) \nonumber +0.1dW_1 \\
    dX_2 &= -\texttt{scale} \bigg(X_1 - X_1^{\text{center}}\bigg)\nonumber+0.1dW_2
\end{align}
This is a standard constant curl representation for a vortex of scale $\texttt{scale}$, centered at \(dX_1, dX_2\), with a potential elliptical deformation determined by $\texttt{logyscale}$. We implement this reference family in practice as a PyTorch \texttt{nn.Module} with four scalar parameters, $X_1^{\text{center}}, X_2^{\text{center}}, \texttt{logyscale}$ and $\texttt{scale}$ to be learned by gradient descent. This module is then used to evaluate \(dX_1, dX_2\), returning \texttt{torch.stack([$dX_1, dX_2$])} in the \texttt{nn.forward} method. The learning process involves optimizing the parameters using gradient descent, with a learning rate of 0.05 over 20 epochs, as determined by the grid search described in \cref{app:inverseproj}.

\subsubsection{Results.}
\label{app:GoM-app-results}

\begin{figure*}[!ht]
    \centering
    \includegraphics[width =0.9\textwidth]{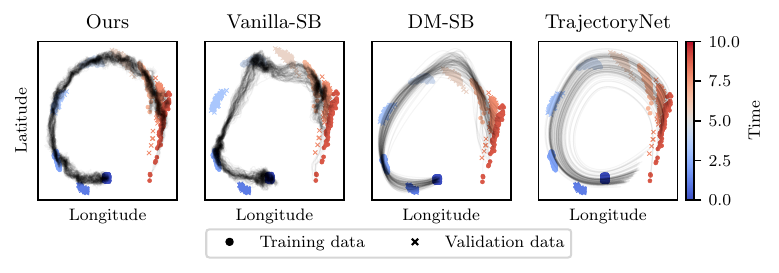}
    \caption{ Comparison on the Gulf of Mexico - big vortex data with 5 training times, 4 validation times, and approximately 111 observations per time. Each plot shows approximately 111 simulated trajectories, originating from particles at one time end point (three left plots: first time; right plot: final time).} 
    \label{fig:GoM-appendix}
\end{figure*}

We provide a visual representation ot the trajectories for the small vortex in \cref{fig:GoMsmall-main} in the main text and for the big vortex in \cref{fig:GoM-appendix}. In terms of EMD and MMD metrics, we show the results for the small vortex in \cref{tab:GoMsmall_EMD_appendix} and \cref{tab:GoMvortexsmall_MMD}, respectively. And for the big vortex in \cref{tab:GoMlarge_EMD_appendix} and \cref{tab:GoMvortex_MMD}. 

For the small vortex, we can see that our method has similar performance to \dmsbname{} on the restricted set of trajectories. And these two methods are both better than the other two alternatives. When considering results over all possible trajectories (last two rows), our method performs even better. If we look at \cref{fig:GoMsmall-main}, we can see that \vanilla{} fails to capture the curvature. \dmsbname{} and TrajectoryNet generate smooth trajectories that are notably far from the data the final validation time point. Our trajectories track the curvature of the validation data closely.

Similar results are observed for the big vortex experiment. Our method has similar performance to \dmsbname{} on the restricted set of trajectories. In particular, according to EMD \dmsbname{} is better for the first time step, similar for the second, and worse for the last two. By comparing the trajectories visually in \cref{fig:GoM-appendix}, we also see that \vanilla{} fails to capture the curvature. \dmsbname{} generates smooth trajectories that fail to capture the third validation time point (and have a slightly unnatural shape in the top right corner of the figure). TrajectoryNet generate smooth trajectories that fail to capture the first validation time point. Our model generates trajectories that most closely resembles the shape that we observe in the ocean current in \cref{fig:gomdata}.

\begin{table*}[!ht]
    \centering
    \begin{tabular}{lcccc}
    \hline
        Method & EMD $t_2$ & EMD $t_4$ & EMD $t_6$ & EMD $t_8$ \\
    \hline
    \vanillaonetime & 0.27 $\pm$ 0.060 & 0.30 $\pm$ 0.056 & 0.43 $\pm$ 0.053 & 0.42 $\pm$ 0.048 \\

         \dmsb & \cellcolor{green!25}\textbf{0.086 $\pm$ 0.011} & \cellcolor{green!25}\textbf{0.092 $\pm$ 0.020} & \cellcolor{green!25}\textbf{0.088 $\pm$ 0.009} & 0.21 $\pm$ 0.027\\
         \trajnet & 0.78  $\pm$ 0.014 & 0.77$\pm$0.048 & 1.19$\pm$0.15 & 0.76$\pm$0.17 \\
         \oursonetime & \cellcolor{green!25}\textbf{0.075 $\pm$ 0.023} & \cellcolor{green!25}\textbf{0.080 $\pm$ 0.017} & 0.13 $\pm$ 0.037 & \cellcolor{green!25}\textbf{0.11 $\pm$ 0.032} \\
    \hline
     \vanillaalltime & 0.27 $\pm$ 0.058 & 0.30 $\pm$ 0.056 & 0.42 $\pm$ 0.056 & 0.41 $\pm$ 0.048 \\
   \oursalltime & \cellcolor{blue!10}\textbf{0.073 $\pm$ 0.020} & \cellcolor{blue!10}\textbf{0.072 $\pm$ 0.012} & 0.12 $\pm$ 0.029 & \cellcolor{blue!10}\textbf{0.094 $\pm$ 0.023} \\
    \hline
    \end{tabular}
    \caption{Earth mover's distance (mean $\pm$ standard deviation) in four validation time points in Gulf of Mexico - small vortex dataset. Results were averaged over 10 seeds.}
    \label{tab:GoMsmall_EMD_appendix}
\end{table*}

\begin{table*}[!ht]
    \centering
    \begin{tabular}{lcccc}
    \hline
        Method & EMD $t_2$ & EMD $t_4$ & EMD $t_6$ & EMD $t_8$ \\
    \hline
     \vanillaonetime & 0.35 $\pm$ 0.043 & 0.55 $\pm$ 0.067 & 0.42 $\pm$ 0.047 & 0.21 $\pm$ 0.038 \\
         
         \dmsb & \cellcolor{green!25}\textbf{0.074 $\pm$ 0.029} & \cellcolor{green!25}\textbf{0.12 $\pm$ 0.027} & 0.19 $\pm$  0.040  & 0.17 $\pm$ 0.048\\
         \trajnet & 1.94  $\pm$ 0.015 & 1.37$\pm$0.043 & 1.27$\pm$0.10 & 1.68$\pm$0.11 \\
         \oursonetime & 0.19 $\pm$ 0.044 & \cellcolor{green!25}\textbf{0.13 $\pm$ 0.038} & \cellcolor{green!25}\textbf{0.092 $\pm$ 0.021} & \cellcolor{green!25}\textbf{0.13 $\pm$ 0.020} \\
    \hline
    \vanillaalltime & 0.35 $\pm$ 0.043 & 0.55 $\pm$ 0.071 & 0.41 $\pm$ 0.050 & 0.20 $\pm$ 0.035 \\
     \oursalltime & 0.20 $\pm$ 0.038 & \cellcolor{blue!10}\textbf{0.13 $\pm$ 0.040} & \cellcolor{blue!10}\textbf{0.086 $\pm$ 0.020} & \cellcolor{blue!10}\textbf{0.13 $\pm$ 0.017} \\
    \hline
    \end{tabular}
    \caption{Earth mover's distance (mean $\pm$ standard deviation) in four validation time points in Gulf of Mexico - big vortex dataset. Results were averaged over 10 seeds.}
    \label{tab:GoMlarge_EMD_appendix}
\end{table*}

\begin{table*}[!ht]
    \centering
    \begin{tabular}{lcccc}
    \hline
        Method & MMD $t_2$ & MMD t=2 & MMD $t_4$ & MMD t=4 \\
    \hline
    \vanillaonetime & 0.73 \fpm 0.32&  0.86\fpm 0.33& 1.67\fpm 0.37 & 1.48\fpm 0.32 \\
         \dmsb & \cellcolor{green!25}\textbf{0.07 $\pm$ 0.02} & 0.08 $\pm$ 0.04 & \cellcolor{green!25}\textbf{0.06 $\pm$ 0.02} & 0.34 $\pm$ 0.11 \\
         \trajnet & 4.38 $\pm$ 0.25 & 4.38 $\pm$ 0.40 & 7.36 $\pm$ 0.76 & 4.16 $\pm$ 1.28 \\
         \oursonetime & \cellcolor{green!25}\textbf{0.042 \fpm 0.035} & \cellcolor{green!25}\textbf{0.043 \fpm 0.031} & 0.16 \fpm 0.10 & \cellcolor{green!25}\textbf{0.064 \fpm 0.031} \\
         
    \hline
    
    \vanillaalltime & 0.72 $\pm$ 0.30 & 0.87 $\pm$ 0.34 & 1.67 $\pm$ 0.39 & 1.44 $\pm$ 0.32 \\

    \oursalltime & \cellcolor{blue!10}\textbf{0.04 $\pm$ 0.03} & \cellcolor{blue!10}\textbf{0.03 $\pm$ 0.02} & 0.14 $\pm$ 0.07 & \cellcolor{blue!10}\textbf{0.05 $\pm$ 0.04} \\
    \hline
    \end{tabular}
    \caption{Maximum mean discrepancy (mean $\pm$ standard deviation) in four validation time points in Gulf of Mexico - small vortex dataset. Results were averaged over 10 seeds.}
    \label{tab:GoMvortexsmall_MMD}
\end{table*}

\begin{table*}[!ht]
    \centering
    \begin{tabular}{lcccc}
    \hline
        Method & MMD $t_2$ & MMD t=2 & MMD $t_4$ & MMD t=4 \\
    \hline
         \vanillaonetime & 1.11 \fpm 0.26& 2.52\fpm  0.53& 1.25\fpm  0.32& 0.27\fpm  0.14 \\
         \dmsb & \cellcolor{green!25}\textbf{0.05 $\pm$ 0.05} & \cellcolor{green!25}\textbf{0.10 $\pm$ 0.07} & \cellcolor{green!25}\textbf{0.30 $\pm$ 0.15} & \cellcolor{green!25}\textbf{0.19 $\pm$ 0.13} \\
         \trajnet & 9.40 $\pm$ 0.04 & 8.19 $\pm$ 0.20 & 7.72 $\pm$ 0.53 & 8.92 $\pm$ 0.17 \\
         \oursonetime & 0.35 \fpm 0.17 & \cellcolor{green!25}\textbf{0.15\fpm  0.10} & \cellcolor{green!25}\textbf{0.42\fpm  0.031} & 0.49\fpm  0.028\\
         \hline
         \vanillaalltime & 1.14 $\pm$ 0.26 & 2.53 $\pm$ 0.55 & 1.50 $\pm$ 0.34 & \cellcolor{blue!10}\textbf{0.27 $\pm$ 0.13} \\
         \oursalltime & 0.37 $\pm$ 0.15 & \cellcolor{blue!10}\textbf{0.13 $\pm$ 0.09} & \cellcolor{blue!10}\textbf{0.04 $\pm$ 0.03} & \cellcolor{blue!10}\textbf{0.06 $\pm$ 0.03} \\
    \hline
    \end{tabular}
    \caption{Maximum mean discrepancy (mean $\pm$ standard deviation) in four validation time points in Gulf of Mexico - big vortex dataset. Results were averaged over 10 seeds.}
    \label{tab:GoMvortex_MMD}
\end{table*}

\clearpage

\subsection{Single cell datasets}
\label{app:eb-app}

\subsubsection{Experiment setup} 
\label{app:singlecellsetup}
We consider two different single cell data for this experiment: the one from \citet{moon2019visualizing} on embryoid body cells (EB) and the one from \citet{chu2016single} on human embryonic stem cells (hESC). Both datasets are shared under CC-by-4.0 license. Both datasets offer insights into the dynamic process of stem cell differentiation by capturing gene expression levels at different stages.
\citet{tong2020trajectorynet} applied a pre-processing pipeline, including dimensionality reduction via Principal Component Analysis, to the EB data. We use the same pre-processing for the EB data; we apply the same pipeline to the hESC data as well. The EB dataset consists of five snapshots that are largely overalapping so we subsampled it to have 300 cells at each snapshot. The hESC dataset initially consists of six snapshots. Note that we want validation data at every other snapshot, and we also want the first and last snapshot to serve as training points. Therefore, we want the total number of snapshots to be an odd number. To meet this desideratum, we choose to ignore the last snapshot of the hESC data. The EB data already meets this desideratum. To apply our method to this dataset, we set $\Delta t = 0.01$ for every discretization step involving SDEs. We initialize the reference drift to 0, so that the initial reference SDE is a simple Brownian motion.

\subsubsection{Reference family choice} 
\label{app:eb_implementation}
For this experiment, we do not have access to the data generating process, so we cannot proceed as done for the synthetic experiments. Nonetheless, scientists have domain knowledge about the underlying dynamics of cell differentiation process. In particular, following \citet{wang2011quantifying, weinreb2018fundamental, lavenant2024toward}, we use a gradient field family. This family is motivated by Waddington's famous analogy between cellular differentiation and a marble rolling down a potential surface \citep{waddington2014strategy}. In practice, we parameterize the gradient field using a multilayer perceptron, an architecture used for gradient field in literature \citep[e.g.,][]{greydanus2019hamiltonian,lin2023computing}. We tested several potential models with one hidden layer of size 128, two hidden layers of size 128 and 64, three hidden layers of sizes 128, 64, and 64, and three hidden layers of sizes 128, 128, and 64 connected by ReLU activation functions, trained with learning rate 0.01 with 20 epochs. We choose based on the criteria described in \cref{app:inverseproj}. The final training hyperparameters is set to 0.01 and we trained for 50 epochs, chosen by the grid search criteria described in \cref{app:inverseproj} after choosing the architecture.

\subsubsection{Results.} 
\begin{figure}[!ht]
    \centering
    \includegraphics[width=\textwidth]{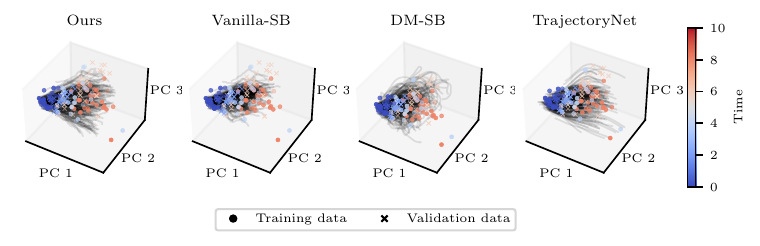}
    \caption{EB dataset}
    \label{fig:eb-app}
\end{figure}

\begin{figure}[!ht]
    \centering
    \includegraphics[width=\textwidth]{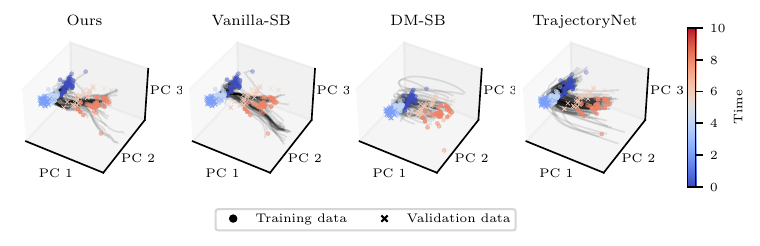}
    \caption{hESC dataset}
    \label{fig:hesc-app}
\end{figure}

In terms of visual reconstruction, \cref{fig:eb-app} shows that all the methods perform well in reconstructing trajectories in the EB experiments. This success can be attributed to the limited number of time steps, which results in restricted global information. Nonetheless, the trajectories for our method, \vanilla{}, and TrajectoryNet look more reasonable at interpolating the data. If we consider instead \cref{fig:hesc-app}, we can see how for this dataset it is very clear that both our method, \vanilla{}, and TrajectoryNet outperform \dmsbname. This approach, indeed, unfortunately seems unable to capture trajectories going through the observations in a meaningful way, whereas the other methods come up with trajectories that are visually reasonable given the observed data. This contradicts what we see in \cref{tab:EB_EMD_appendix} and \cref{tab:EB_MMD}. Indeed, when comparing the quality of only those trajectories generated from a single end point in time, \dmsbname{} outperforms the alternatives, including our method, for both EMD and MMD. However, when allowed to generate trajectories from all particles, \vanilla{} and our method outperform the single-time trajectory options (as expected) and perform comparably to each other.

\begin{table*}[!ht]
    \centering
    \begin{tabular}{lcccc}
    \hline
       & \multicolumn{2}{c}{EB} & \multicolumn{2}{c}{hESC} \\
        Method & EMD $t_2$ & EMD $t_4$ & EMD $t_2$ & EMD $t_4$ \\
    \hline
        \vanillaonetime & 1.49 $\pm$ 0.063 & 1.55 $\pm$ 0.034 & 1.47 $\pm$ 0.088 & 1.97 $\pm$ 0.169\\
         \dmsb & \cellcolor{green!25}\textbf{1.13 $\pm$ 0.082} & \cellcolor{green!25}\textbf{1.45 $\pm$ 0.16} & \cellcolor{green!25}\textbf{1.10 $\pm$ 0.066} & 1.51 $\pm$ 0.11\\
         \trajnet & 2.03 $\pm$ 0.04 & 1.93 $\pm$ 0.08 & 1.30 $\pm$ 0.04 & 1.93 $\pm$ 0.05 \\
         \oursonetime & 1.27 $\pm$ 0.028 & 1.57 $\pm$ 0.048 & \cellcolor{green!25}\textbf{1.08 $\pm$ 0.12} & \cellcolor{green!25}\textbf{1.33 $\pm$ 0.084} \\
    \hline
        \vanillaalltime & \cellcolor{blue!10}\textbf{1.12 $\pm$ 0.031} & \cellcolor{blue!10}\textbf{1.12 $\pm$ 0.023} & \cellcolor{blue!10}\textbf{0.72 $\pm$ 0.017} & \cellcolor{blue!10}\textbf{1.27 $\pm$ 0.043} \\
        \oursalltime & \cellcolor{blue!10}\textbf{0.96 $\pm$ 0.019} & \cellcolor{blue!10}\textbf{1.19 $\pm$ 0.017} & \cellcolor{blue!10}\textbf{0.71 $\pm$ 0.031} & \cellcolor{blue!10}\textbf{1.25 $\pm$ 0.076} \\
    \hline
    \end{tabular}
    \caption{Earth mover's distance (mean $\pm$ standard deviation) in two validation time points in EB and hESC datasets.}
    \label{tab:EB_EMD_appendix}
\end{table*}


\begin{table*}[!ht]
    \centering
    \begin{tabular}{lcccc}
    \hline
       & \multicolumn{2}{c}{EB} & \multicolumn{2}{c}{hESC} \\
        Method & MMD $t_2$ & MMD $t_4$ & MMD $t_2$ & MMD $t_4$ \\
    \hline
        \vanillaonetime & 0.68 \fpm 0.035  & 0.65 \fpm  0.066 & 5.23 \fpm  0.24  & 5.19  \fpm  0.22 \\
         
         \dmsb & \cellcolor{green!25}\textbf{0.24 $\pm$ 0.035} & \cellcolor{green!25}\textbf{0.19 $\pm$ 0.038} & \cellcolor{green!25}\textbf{3.22 $\pm$ 0.21} & \cellcolor{green!25}\textbf{3.23 $\pm$ 0.27} \\
         \trajnet & 0.62 $\pm$ 0.02 & 0.60 $\pm$ 0.10 & 4.03 $\pm$ 0.16 & 4.42 $\pm$ 0.17 \\
         \oursonetime & 0.45 \fpm 0.022 & 0.54 \fpm 0.081 & 3.71 \fpm 0.43 & \cellcolor{green!25}\textbf{3.49 \fpm 0.50} \\
    \hline
        \vanillaalltime & \cellcolor{blue!10}\textbf{0.19 $\pm$ 0.021} & \cellcolor{blue!10}\textbf{0.19 $\pm$ 0.017} & \cellcolor{blue!10}\textbf{1.85 $\pm$ 0.11} & \cellcolor{blue!10}\textbf{2.77 $\pm$ 0.17} \\
        \oursalltime & \cellcolor{blue!10}\textbf{0.14 $\pm$ 0.015} & \cellcolor{blue!10}\textbf{0.16 $\pm$ 0.011} & \cellcolor{blue!10}\textbf{2.02 $\pm$ 0.19} & \cellcolor{blue!10}\textbf{3.35 $\pm$ 0.30} \\
    \hline
    \end{tabular}
    \caption{Maximum mean discrepancy (mean $\pm$ standard deviation) in two validation time points in EB and hESC datasets.}
    \label{tab:EB_MMD}
\end{table*}

\clearpage

\subsection{Convergence Analysis of Test Metrics Across Iterations}
\label{app:multisteps-app}
\begin{figure}[!h]
    \centering
    \includegraphics[width=\linewidth]{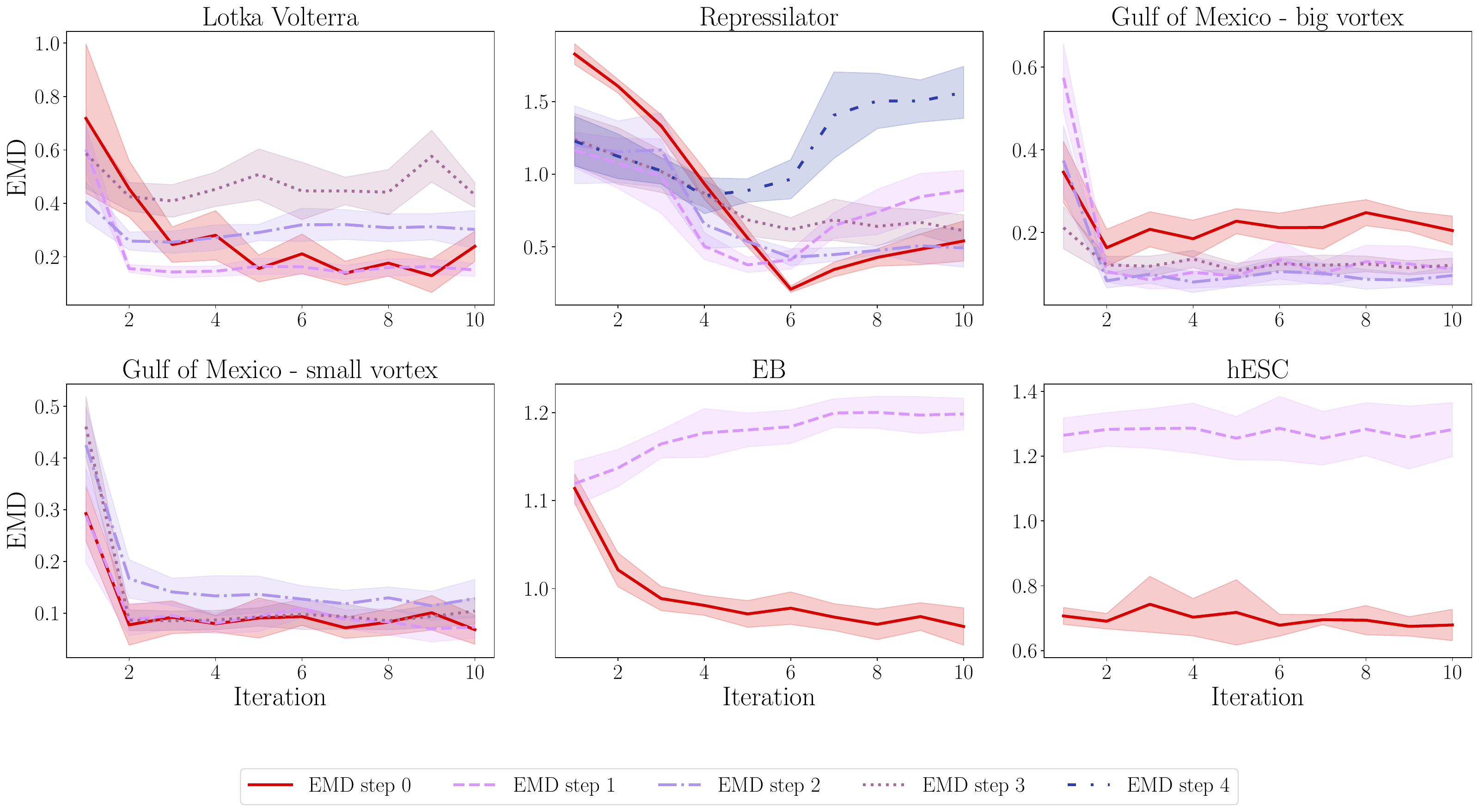}
    \caption{Evolution of the EMD metric across ten iterative refinement steps in our proposed method for different experimental settings. Each subplot corresponds to a specific experiment: Lotka-Volterra, Repressilator, Gulf of Mexico (big vortex), Gulf of Mexico (small vortex), EB, and hESC. The x-axis represents the iteration number, ranging from 1 to 10, while the y-axis indicates the EMD value at each validation time point. The different curves within each subplot correspond to the EMD measured at multiple validation steps.}
    \label{fig:emditer}
\end{figure}

\begin{figure}[!h]
    \centering
    \includegraphics[width=\linewidth]{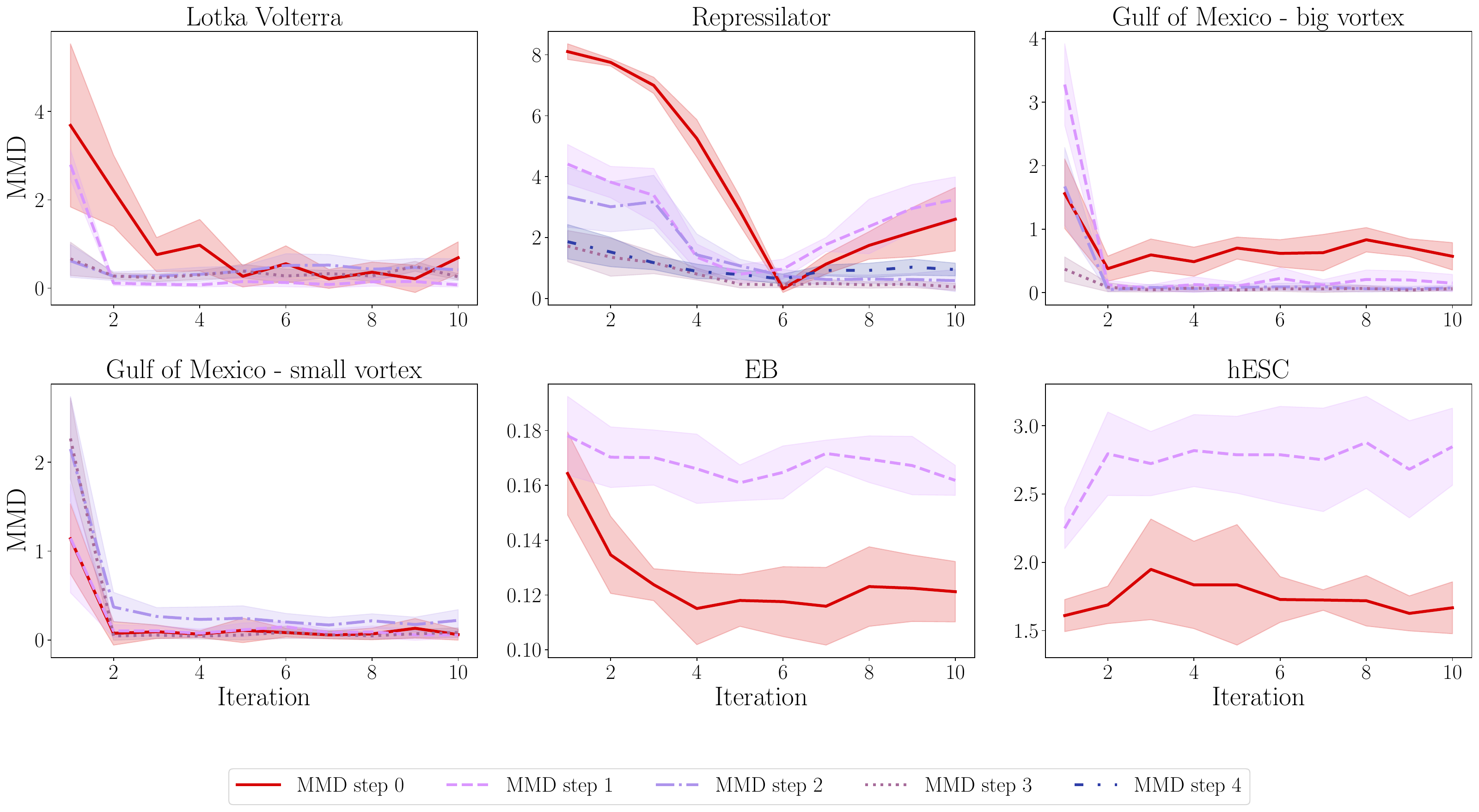}
    \caption{Evolution of the MMD metric across ten iterative refinement steps in our proposed method for different experimental settings. Each subplot corresponds to a specific experiment: Lotka-Volterra, Repressilator, Gulf of Mexico (big vortex), Gulf of Mexico (small vortex), EB, and hESC. The x-axis represents the iteration number, ranging from 1 to 10, while the y-axis indicates the EMD value at each validation time point. The different curves within each subplot correspond to the EMD measured at multiple validation steps.}
    \label{fig:mmditer}
\end{figure}

In this section, we present the evolution of the test metrics --- MMD and EMD --- over ten iterative reference steps in our proposed method. Figures \ref{fig:emditer} and \ref{fig:mmditer} illustrate how these metrics change across different experimental settings. Specifically, \ref{fig:emditer} depicts the EMD at various validation time points, while \ref{fig:mmditer} displays the MMD trends over the same iterations. Our observations indicate that in most cases, the method exhibits initial improvements within the first few iterations, leading to a decrease in both EMD and MMD. Typically, ten iterations are sufficient for the refinement steps to reach stability, confirming the effectiveness of our iterative approach. However, in some validation steps --- such as the first two steps in the Repressilator experiment --- an increase in the test metric is observed after initial improvements, suggesting possible overfitting. This phenomenon highlights the importance of developing principled stopping criteria to prevent performance degradation beyond a certain number of iterations. These results provide insights into the iterative refinement behavior of our method and suggest potential directions for future work in adaptive stopping mechanisms.

\clearpage

\section{Different Baselines Timing}
\label{app:timing-table}

In this section we provide a table reporting the computing times of the four methods when tested on the 6 tasks of interest. The experiments were run on four cores of Intel Xeon Gold 6248 CPU and one Nvidia Volta V100 GPU with 32 GB RAM.  

In \cref{tab:timing-tables-raw} we report the raw times for all the methods over the various tasks. The times are reported in \textbf{hours}. The format is mean $\pm$ standard deviation, where these quantities are obtained by evaluating computing times over 10 runs with 10 different seeds. We observe that \vanilla{} is the fastest method, with all tasks finishing in less than 30 minutes, and as little as few minutes for Lotka Volterra, EB, and hESC. Our method is also efficient. In general, it takes approximately 10 times longer than \vanilla{}, and this is because (1) we have to solve an SB problem $K=10$ times, and (2) the \texttt{MLEFit} step is much faster than the SB estimation itself. \dmsbname{} and TrajectoryNet are much slower. \dmsbname{} takes an average of 15 hours across all experiments, with only Lotka Volterra being a bit better (but still approximately 8 hours). TrajectoryNet takes a total of 7 to 11 hours across the various tasks. This is much slower when compared to our own method. 

To further analyze how faster our method is, we provide in \cref{tab:timing-tables-ratios} the ratios between alternative algorithms and our method's elapsed times over the different tasks. Also for this table, we provide average and standard deviation across 10 seeds. We can immediately notice that \vanilla{} is 10 times faster, as expected. \dmsbname{} ranges from being 3 to 41 times slower, with an overall average of 16 times slower across all tasks (see \cref{tab:timing-avg-avg}). TrajectoryNet performs better than \dmsbname{} but remains slower than our method, with ratios ranging from approximately 2 to 27 times slower, averaging 11 times slower overall. These results show that our method not only achieves better trajectory inference results in most experiments, but it is also much faster than all the alternatives that nontrivially handle multiple time points. We include task-specific runtime discussions in the ``Runtime" paragraphs for each experiment in \cref{sec:experiments-main}.

\begin{table}[!ht]
    \centering
\begin{tabular}{|l|l|r|r|}
\toprule
algorithm & task &  mean & std \\
\midrule
\multirow[t]{9}{*}{\dmsbname} & Lotka Volterra & 7.62 & 3.18 \\
 & Repressilator & 15.63 & 0.12 \\
 & Gulf of Mexico - big vortex & 15.54 & 0.23 \\
 & Gulf of Mexico - small vortex & 15.44 & 0.02 \\
 & EB & 15.54 & 0.41 \\
 & hESC & 15.40 & 0.08 \\
 
\cline{1-4}
\multirow[t]{9}{*}{\vanilla{}} & Lotka Volterra & 0.06 & 0.01 \\
 & Repressilator & 0.23 & 0.05 \\
 & Gulf of Mexico - big vortex & 0.44 & 0.05 \\
 & Gulf of Mexico - small vortex & 0.43 & 0.01 \\
 & EB & 0.03 & $<$0.01 \\
 & hESC & 0.05 & $<$0.01 \\
 
\cline{1-4}
\multirow[t]{9}{*}{TrajectoryNet} & Lotka Volterra & 10.96 & 0.81 \\
 & Repressilator & 9.86 & 0.43 \\
 & Gulf of Mexico - big vortex & 8.10 & 0.28 \\
 & Gulf of Mexico - small vortex & 7.44 & 0.25 \\
 & EB & 10.19 & 0.37 \\
 & hESC & 8.00 & 0.49 \\
 
\cline{1-4}
\multirow[t]{9}{*}{Ours} & Lotka Volterra & 0.61 & 0.12 \\
 & Repressilator & 2.43 & 0.60 \\
 & Gulf of Mexico - big vortex & 4.68 & 0.75 \\
 & Gulf of Mexico - small vortex & 4.67 & 0.66 \\
 & EB & 0.38 & 0.05 \\
 & hESC & 0.56 & 0.04 \\
 
\bottomrule
\end{tabular}

\caption{Tables with raw timing in \textbf{hours} aggregated over 10 seeds.}
    \label{tab:timing-tables-raw}
\end{table}

\begin{table}[!ht]
    \centering
\begin{tabular}{|l|l|r|r|}
\toprule
algorithm & task &  mean & std \\
\midrule
\multirow[t]{9}{*}{\dmsbname} & Lotka Volterra & 11.98 & 6.28 \\
 & Repressilator & 6.71 & 1.25 \\
 & Gulf of Mexico - big vortex & 3.39 & 0.45 \\
 & Gulf of Mexico - small vortex & 3.36 & 0.39 \\
 & EB & 41.17 & 4.80 \\
 & hESC & 27.43 & 1.96 \\
 
\cline{1-4}
\multirow[t]{9}{*}{\vanilla{}} & Lotka Volterra & 0.10 & 0.01 \\
 & Repressilator & 0.10 & 0.01 \\
 & Gulf of Mexico - big vortex & 0.10 & 0.01 \\
 & Gulf of Mexico - small vortex & 0.09 & 0.01 \\
 & EB & 0.09 & 0.01 \\
 & hESC & 0.09 & 0.01 \\
 
\cline{1-4}
\multirow[t]{9}{*}{TrajectoryNet} & Lotka Volterra & 18.37 & 3.29 \\
 & Repressilator & 4.25 & 0.87 \\
 & Gulf of Mexico - big vortex & 1.77 & 0.25 \\
 & Gulf of Mexico - small vortex & 1.62 & 0.21 \\
 & EB & 26.97 & 2.82 \\
 & hESC & 14.26 & 1.28 \\
 
\cline{1-4}
\bottomrule
\end{tabular}

\caption{Table with the fold of acceleration of our method against alternatives (rounded to two decimal places). To compute each entry we first evaluate, for each (seed, algorithm, task) tuple, the ratio between the elapsed time for that experiment and the elapsed time for the same task and seed when our model is run. We then average these ratios across the 10 seeds and report mean and standard deviation.}
    \label{tab:timing-tables-ratios}
\end{table}

\begin{table}[!ht]
    \centering
    \begin{tabular}{|l|r|r|}
\toprule
algorithm &  mean &  std \\
\hline
                \dmsbname &        16.08 &       15.13 \\
               \vanilla{} &       0.09 &       0.01 \\
             TrajectoryNet &        11.21 &       9.72 \\
\hline
\bottomrule
\end{tabular}
    \caption{In this table, we report --- for each alternative algorithm --- average ratio speedups across all seeds and tasks. To compute this, we compute the ratios as explained in the caption of \cref{tab:timing-tables-ratios}, and then for each alternative algorithm we average these ratios over the 10 seeds and 6 tasks.}
    \label{tab:timing-avg-avg}
\end{table}

\clearpage

\section{Computational Limitations for \dmsbname\ and TrajectoryNet}
\label{app:comp-challenges}

TrajectoryNet uses continuous normalizing flows (CNFs) to model continuous-time dynamics. While CNFs are powerful for modeling complex distributions, they are often computationally intensive. Several studies have highlighted the computational challenges associated with CNFs. For example, \citet{grathwohl2018ffjord} discuss the overhead of integrating neural networks in CNFs and propose methods like FFJORD to improve efficiency. One of the key issues is that to train these methods the algorithms need to compute the trace of the Jacobian of the transformation function at each iteration. This operation is computationally expensive, especially in high-dimensional spaces \citep{chen2018neural}. The computational burden increases when regularization terms are added, as in TrajectoryNet, to enforce desired properties on the flow, further slowing down the training process.

\dmsbname{} goes beyond standard SB frameworks by tailoring the Bregman
Iteration and extending the Iteration Proportional Fitting algorithm to phase space. 
Rather than modeling particles' locations directly, this approach augments the observations with random velocities, modeling particles’ velocity and location jointly through a Langevin dynamic. We hypothesize that \dmsbname{} faces computational challenges due to learning dynamics in this expanded space (velocity and location versus location alone), as it deals with (1) the absence of direct observations for the velocity component and (2) potential inconsistencies between randomly sampled initial velocities and observed location changes (i.e., the predicted new locations using the sampled velocities may not align with the observed locations). In the context of generative modeling using diffusion, \citet{dockhorn2021score} also augment their model with particle velocities and face increased training time.

\section{Identifiability concerns}
\label{sec:identifiability}

In some scenarios, observing marginal samples may not provide enough information to fully understand the underlying system. One situation where this can occur is when the system starts in equilibrium, such as when the initial distribution is the system's invariant measure (if one exists). However, this lack of information can also arise even when the system is not in equilibrium or when no invariant measure is present. For example, consider a system where the drift consists of a rotationally symmetric gradient field combined with a constant-curl rotational component. If the initial sample distribution is also rotationally symmetric, the marginal distribution would retain this symmetry at every time step, regardless of the angular velocity introduced by the constant-curl rotation. As a result, by only observing marginal samples, we might miss the rotational component that governs the angular velocity, preventing us from accurately inferring the system's trajectories.

\paragraph{A specific example of a rotationally invariant vector field.} Consider a vector field with constant curl, starting from an isotropic Gaussian distribution. Formally, in the context of \cref{eq:mainsde}, consider a simple two-dimensional drift $\drift(\bm{x}) = [\alpha x_2, -\alpha x_1]^\top$,
where \(\alpha \in \mathbb{R}\) is a parameter. This represents the dynamics of a vector field with constant curl. When \(\alpha = 0\), the system is purely driven by Brownian motion. Then assume the initial distribution \(\marginal{0}\) is isotropic normal, i.e., \(\marginal{0} \sim \mathcal{N}(0, \beta I_2)\) for some \(\beta\), where \(I_2\) is the 2D identity matrix. For a fixed volatility \(\gamma\), when \(\alpha = 0\), the particle distribution at any future time step remains an isotropic normal distribution.

Indeed, if we consider the general form of the Fokker--Planck equation,
\begin{align*}
    \frac{\partial p(\bm x, t)}{\partial t}
    = -\nabla\cdot \bigl(\drift(\bm x)\,p(\bm x, t)\bigr)
      + \gamma \nabla^2 p(\bm x, t)
    = - p(\bm x, t) \,\bigl(\nabla\cdot \drift(\bm x)\bigr)
      - \nabla p(\bm x, t)\cdot \drift(\bm x)
      + \gamma \nabla^2 p(\bm x, t),
\end{align*}
we see that for this isotropic Gaussian setting, \(\nabla p(\bm x, t)\cdot \drift(\bm x)=0\). Hence only the diffusion term remains, and the distribution of the particles at any time step stays isotropic Gaussian with increasing variance. Schrödinger bridges between these Gaussians have a unique closed-form \citep{Bunne2023}, which cannot account for all possible rotations. This illustrates how certain symmetries (like rotational invariance) can lead to multiple vector fields that produce indistinguishable marginal or trajectory behaviors under certain conditions.

\paragraph{Two Types of Identifiability Concerns.} The above example underscores the broader issue that different underlying systems can give rise to the same observed data. We identify at least two distinct scenarios:

\begin{itemize}
    \item \textbf{(1) Different systems yielding the \emph{same} trajectory.} Two (or more) truly distinct underlying dynamics might coincide on the \emph{exact} same trajectory over time. In this situation, identifying which system actually generated the data is strictly harder (and in some cases, impossible) given only trajectory observations. However, note that \emph{identifying} the system is a different goal from simply producing accurate predictions or interpolations. It is possible to infer a usable trajectory even when the true system remains ambiguous.
    \item \textbf{(2) Different dynamics yielding the same marginal distributions at observed time points, but differing in-between.}
    Even if one has marginals at a few discrete time points, different dynamics can ``fill in'' between these marginals in different ways, leading to distinct trajectories that match the same observed endpoints. This poses a fundamental limitation for trajectory inference unless additional information (e.g., domain knowledge or more finely sampled data) is introduced. Providing uncertainty quantification over all plausible trajectories could be an interesting way to reflect this ambiguity.
\end{itemize}

These concerns mirror those in inverse problems and PDE identifiability. For instance, \citet{lavenant2024toward} showed that if one has marginals sampled arbitrarily densely in time (and the vector field is a gradient field), then the trajectories can indeed be uniquely identified. \citet{guan2024identifying} provide conditions under which a system with \emph{linear drift} and \emph{additive diffusion} can be identified from snapshot data, specifically when the initial distribution is not invariant to a class of generalized rotations. These results demonstrate that identifiability can hold under certain conditions but also indicate how quickly it may fail under symmetries or insufficient sampling.

\paragraph{Relation to Convergence in \Cref{prop:iter-kl}}

In \cref{prop:iter-kl}, we show that the KL divergence in our iterative method is guaranteed to converge because the sequence it forms is a lower-bounded, monotonically decreasing sequence. However, this does not imply that the \emph{arguments} of the KL (i.e., the pair \((p,q)\)) necessarily converge to a \emph{unique} solution. In principle, there could be multiple \((p,q)\) pairs that produce the same KL value and are equally consistent with the observed data. In other words, this is precisely an \emph{identifiability issue}: if multiple solutions fit the same marginal distributions (and/or constraints) equally well, the algorithm might either settle on different solutions depending on initialization or potentially ``cycle'' among equivalent solutions. While we
do not observe such cycling in our experiments, it remains an open theoretical question whether this cycling could occur. Ensuring uniqueness of the solution beyond the decreasing KL criterion may demand additional constraints, richer data sampling, or stronger prior assumptions. We believe investigating these settings, and the broader issue of identifiability, is a fruitful direction for future research.

\end{document}